\documentclass{article} 
\usepackage{amsmath, amsthm, amssymb} 
\usepackage{subcaption}
\usepackage{changepage} 

\usepackage{graphicx}
\usepackage{tikz-cd}
\usepackage{bbm}
\usepackage[margin=1in]{geometry} 
\usepackage{microtype} 
\usepackage{parskip}   

\usepackage[colorinlistoftodos,prependcaption,textsize=small]{todonotes}

\usepackage[backend=biber, style=alphabetic]{biblatex}  
\theoremstyle{definition}

\theoremstyle{remark}

\usepackage[colorlinks=true, linkcolor=red, citecolor=blue, urlcolor=blue]{hyperref}

\title{An Extended Topological Model For High-Contrast Optical Flow}
\author{Brad Turow, Jose A. Perea \thanks{ This work was partially supported by the National Science Foundation through   CAREER award \# DMS-2415445.}}

\begin{document}

\maketitle

\begin{abstract}

In this paper, we  identify low-dimensional models for dense core subsets in the space of $3\times 3$ high-contrast optical flow patches sampled from the Sintel dataset. 
In particular, we leverage the theory of approximate and discrete circle bundles to identify a 3-manifold whose boundary is a previously proposed optical flow torus, together with disjoint circles corresponding to pairs of binary step-edge range image patches. 
The 3-manifold model we introduce provides an explanation for why the previously-proposed torus model could not be verified with direct methods (e.g., a straightforward persistent homology computation). 
We also demonstrate that nearly all optical flow patches in the top 1 percent by contrast norm are found near the family of binary step-edge circles described above, rather than the optical flow torus, and that these frequently occurring patches  are concentrated near motion boundaries (which are of particular importance for computer vision tasks such as object segmentation and tracking). 
Our findings offer   insights on the subtle interplay between topology and geometry in   inference for visual data. \\   
\end{abstract}

\begin{figure}[h]
    \centering
    \begin{minipage}{0.49\textwidth}
        \includegraphics[width=\linewidth]{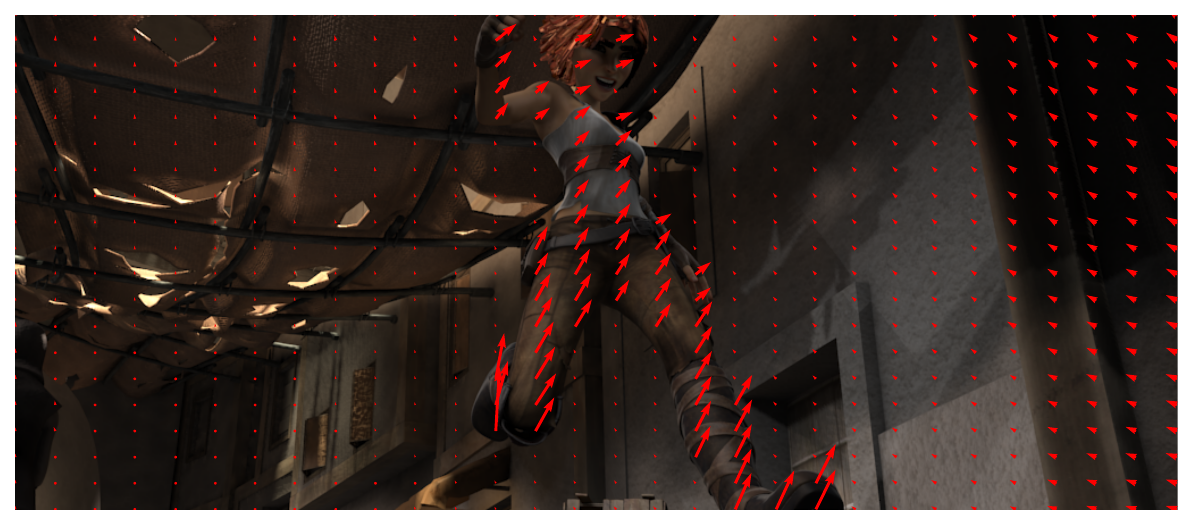}
    \end{minipage}%
    \hfill
    \begin{minipage}{0.49\textwidth}
        \includegraphics[width=\linewidth]{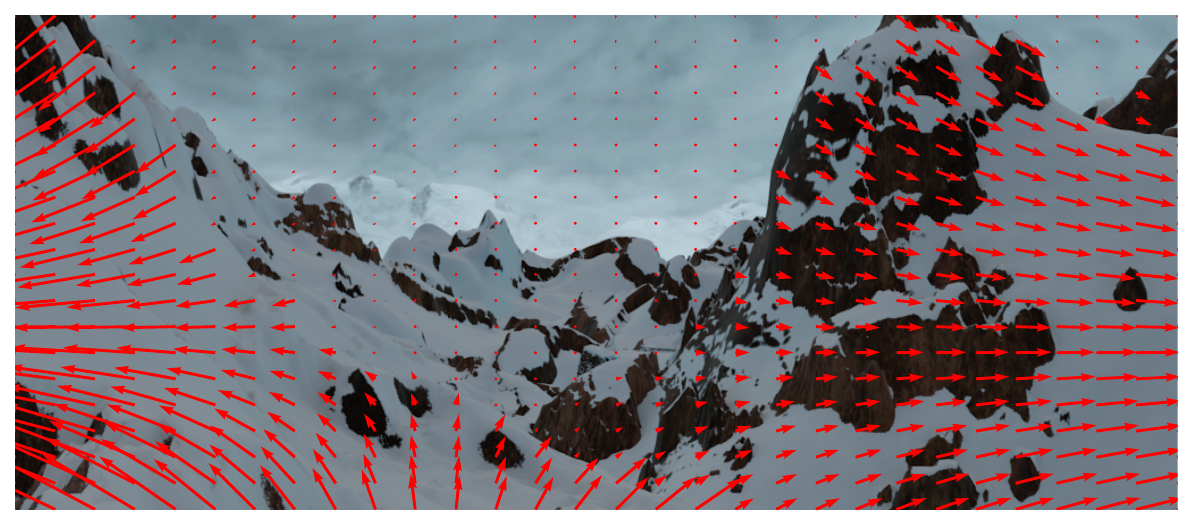}
    \end{minipage}
    \caption{Frames from the Sintel video labeled with sample (scaled) optical flow vectors}
    \label{fig: FramesSintel}
\end{figure}

\section{Motivation And Context}\label{sec: Motivation}
Optical flow describes the perceived motion of objects across successive frames in a video~\cite{Horn_Opt_Flow}. 
This motion can be represented by a two-dimensional vector $v\in\mathbb{R}^{2}$ at each pixel location, indicating where the pixel value at coordinates $(x,y)$ in frame $t$ appears in frame $t+1$. 
Accurately estimating optical flow is a fundamental step in numerous computer vision applications, such as tracking moving objects, segmenting scenes, and compressing video data~\cite{Autonomous_Vehicles, Opt_Flow_Applications}.\\

\noindent However, accurately modeling and analyzing the structure of optical flow data remains a significant challenge~\cite{Opt_Flow_Review}. 
One fundamental difficulty is that optical flow cannot always be directly recovered from video data due to inherent ambiguities such as the ``aperture problem'', where local motion is consistent with multiple possible global motions. 
Additionally, real-world videos often feature complexities such as occlusions, motion blur and non-rigid deformations, which further complicate optical flow estimation~\cite{Opt_Flow_Survey, Sintel_Paper}. \\

\noindent To address these challenges, considerable effort has been dedicated to modeling statistical distributions of optical flow patches~\cite{Opt_Flow_Patch_Detect, Opt_Flow_Patch_Learning, opt_flow_torus}. 
A guiding principle in these efforts is the manifold hypothesis, which posits that high-dimensional data is not uniformly distributed in its ambient space, but instead concentrates around dense subspaces that are well-modeled by low-dimensional manifolds~\cite{Manifold_Hypothesis}. 
When these manifolds exhibit non-trivial topology, tools from topological data analysis (TDA), such as persistent homology, can offer invaluable guiding insights~\cite{TDA_Overview}. \\

\noindent The Sintel dataset~\cite{Sintel}, derived from the open-source animated short film ``Sintel'' by the Blender Foundation, provides a high-quality benchmark for testing and training optical flow models (see Figure \ref{fig: FramesSintel}). 
Its rich variety of motion patterns, coupled with the availability of ground truth optical flow, makes it ideal for studying the statistical and geometric properties of optical flow data ~\cite{Sintel_Paper}. 
In 2020, Adams et al. theorized and provided evidence that the set of $3\times 3$ high-contrast optical flow patches from the Sintel dataset contains a dense core subset which is well-approximated by a space with the topology of a 2-dimensional torus, effectively offering a 2-dimensional model for 18-dimensional data ~\cite{opt_flow_torus}.\\

\noindent In this paper, we confirm and derive coordinates for the proposed torus model as part of a larger 3-manifold structure, and identify another key family of dense core subsets in the space of high-contrast $3\times 3$ optical flow patches from~\cite{Sintel}.
We show that these new subsets correspond to patches concentrated near motion boundaries, which are of special significance for computer vision applications such as motion segmentation, object tracking and boundary detection ~\cite{Motion_Boundary_Detection2,Motion_Boundary_Detection}.\\

\section{Prior Work}\label{sec: Prior Work}

\noindent The analysis of dense core subsets within spaces of optical image, range and flow patches for the purpose of studying nonlinear statistics has a rich history over the past two decades. 
This section provides a brief review of the developments which are directly relevant to our own work.\\

\begin{figure}[h!]
    \centering
    \includegraphics[width = 0.5\textwidth]{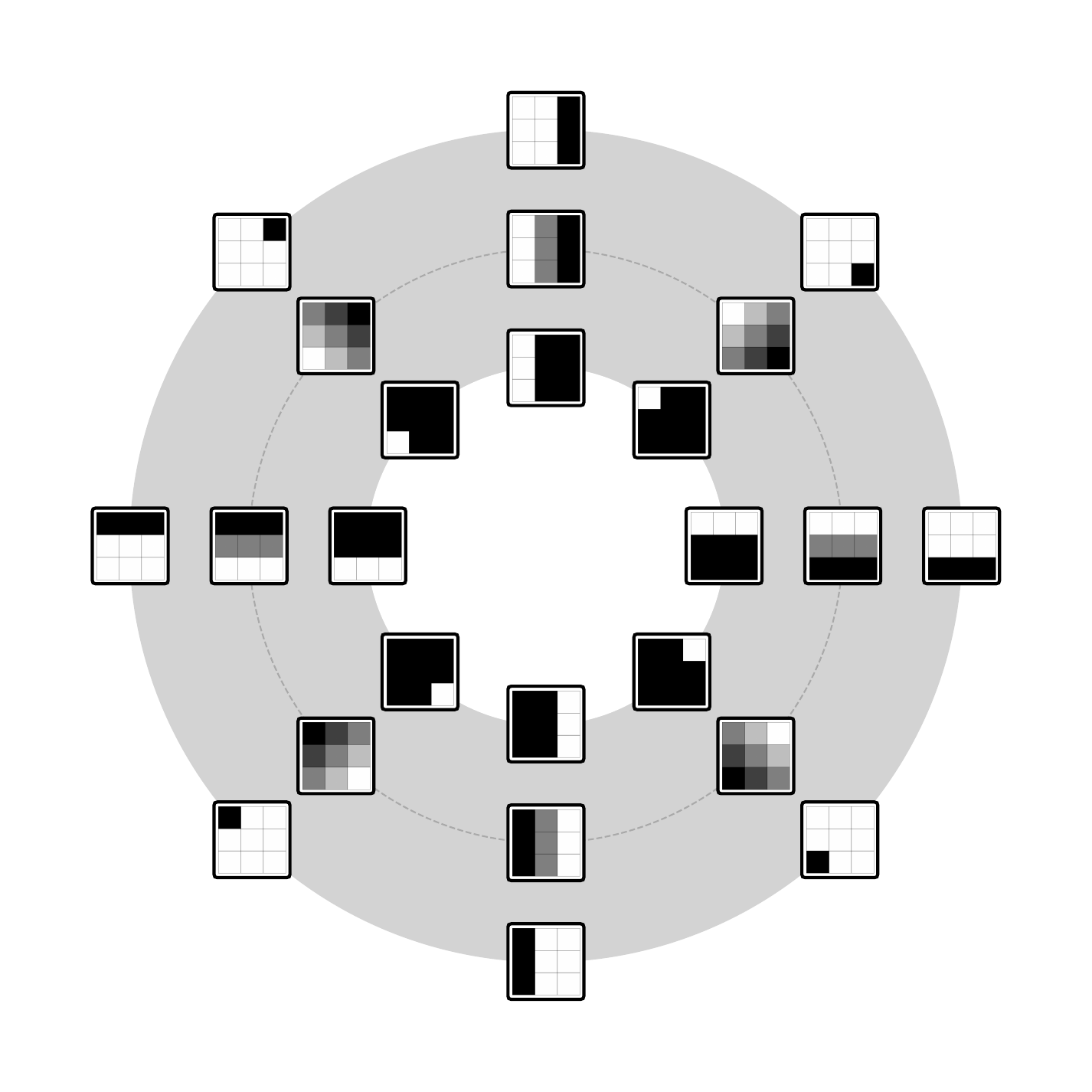}
    \caption{The linear step-edge annulus model for high-contrast $3 \times 3$ optical image patches from \cite{Lee_Mumford}. Some binary step-edge patches are shown on the boundary circles. The central circle is the primary circle described in \cite{Klein_bottle}}
    \label{Range_Step_Annulus}
\end{figure}

\noindent In a seminal paper \cite{Lee_Mumford}, Lee et al. studied the distributions of high-contrast $3 \times 3$ optical range and grayscale image patches sampled from the van Hateren natural image dataset \cite{Van_Hateren_Dataset}. 
They showed that (after mean-centering and contrast normalization) frequently occurring image patches were concentrated around a non-linear 2-dimensional manifold $\mathcal{M}$ of linear step-edges, homeomorphic to an annulus (see Figure~\ref{Range_Step_Annulus}). 
The range patches, on the other hand, were found to be concentrated in distinct clusters around the binary patches in $\mathcal{M}$.\\

\noindent In a subsequent series of papers \cite{Range_Patches, Horizontal_Flow_Circle, Klein_bottle, opt_flow_torus},  topologically-flavored tools such as persistent homology \cite{TDA_Overview} were used to model dense core subsets in spaces of high-contrast image, range and optical flow patches. In \cite{Range_Patches}, they showed that the apparent fundamental differences between the two distributions identified in \cite{Lee_Mumford} dissipate with scale -- in particular, that $5 \times 5 $ and $7 \times 7$ range patches from \cite{Mumford_Dataset} are distributed along continuous manifolds with circular topology, much like the optical image patches. \\

\noindent In \cite{Klein_bottle}, the authors used persistent homology to identify a dense core subset with circular topology in the space of $3 \times 3 $ optical image patches from \cite{Mumford_Dataset}. 
This space, which they called the \textit{primary circle}, is a deformation retract of the annulus model described in \cite{Lee_Mumford}, and consists of those linear step-edge patches for which the edge passes through the center of the patch. Using a finer density estimator, the authors also identified a core subset of optical image patches with the topology of a Klein bottle, containing the primary circle as a subset. 
In~\cite{perea2014klein}, Perea and Carlsson leveraged the Klein bottle model described in~\cite{Lee_Mumford} to develop a texture classification algorithm. \\   

\noindent In \cite{Horizontal_Flow_Circle}, the authors used the nudged elastic-band method to identify a dense core subset with circular topology in the space of $3 \times 3$ high-contrast optical flow patches from \cite{Sintel}. 
This space, which they called the \textit{horizontal flow circle}, is obtained by applying a uniform horizontal camera motion to each patch in the primary circle model from \cite{Klein_bottle}. 
In \cite{opt_flow_torus}, the authors propose that the high-contrast optical flow patches actually occupy an entire torus, with a copy of the horizontal flow circle for each possible direction of camera motion (see Section~\ref{sec: Torus Model}).  \\

\section{Our Contributions}\label{sec: Contributions}

In this paper, we use tools from algebraic topology and topological data analysis, specifically the theory of approximate and discrete circle bundles \cite{turow2025discrete}, to confirm and expand upon the models of \cite{Horizontal_Flow_Circle, opt_flow_torus} for the space of $3 \times 3$ high-contrast optical flow patches sampled from the Sintel dataset \cite{Sintel}. 
We do so in several directions: \\

\noindent In Section~\ref{sec: Predom Dir Map}, we show that the feature map used in \cite{opt_flow_torus} to provide support for the torus model is highly non-Lipschitz (and not well-defined) for a substantial portion of the dense core subset which the model fails to account for. 
This ``extra data" is comprised of high-contrast patches with no clear predominant axis of flow. 
In Section~\ref{sec: Extended Model}, we propose a model extending the dense core subset identified in \cite{opt_flow_torus}: a 3-manifold whose boundary is the optical flow torus. 
We verify the model in Section~\ref{sec: Extended Model Evidence} using the sparse circular coordinates algorithm \cite{Sparse_CC} to locally parameterize the data, and circle bundle inference techniques from~\cite{turow2025discrete} to assemble the global structure. 
The extended model, as the authors of~\cite{opt_flow_torus} indicate,  explains why  the torus model cannot be certified with a direct persistent homology computation. \\

\noindent In Section~\ref{sec: Filaments}, we use a  finer density estimator to identify another important family of dense core subsets in the set of $3 \times 3$ high-contrast optical flow patches from the Sintel dataset~\cite{Sintel}, closely related to the binary step-edge range patches explored in~\cite{Lee_Mumford}.
In Section~\ref{sec: Filaments Results}, we show that the feature map from~\cite{opt_flow_torus} is well-defined on the patches in these subsets. 
We then apply a Mapper-like pipeline~\cite{Mapper} to resolve the data into distinct circular components and use the sparse circular coordinates algorithm from~\cite{Sparse_CC} to show that each component corresponds to a pair of binary step-edge range patches. \\

\noindent In Section~\ref{sec: Filament Interpretation}, we show that nearly all of the optical flow patches in the top 1 percent by contrast norm are found near the binary step-edge circles rather than the optical flow torus. We demonstrate empirically that these exceptional patches appear at motion boundaries, whereas other patches in the top 20 percent (the threshold for high-contrast chosen in \cite{opt_flow_torus}) tend to also appear on the interiors of textured moving bodies such as hair. By analogy with \cite{Range_Patches}, we naturally hypothesize that for optical flow patches of larger size, the torus and binary step-edge circles give rise to a single connected manifold structure -- a parameterized family of linear step-edge annuli like the one shown in Figure~\ref{Range_Step_Annulus} which deformation retracts onto the optical flow torus. \\

\section{Methods}\label{sec: Methods}

Our approach to modeling dense core subsets relies  on tools and ideas from topological data analysis (TDA), briefly summarized below: \\

\noindent\textbf{Persistent Homology: }The primary computational tool in TDA is persistent homology  \cite{perea2019brief}. 
Roughly speaking, given a subset $X$ of a metric space $M$ (e.g., $\mathbb{R}^{D}$), one can consider thickenings of $X$ of the form $X^{(s)} = \bigcup\limits_{x\in X}B_{s}(x)$, where $B_s(x)$ denotes the open ball of radius $s \geq 0$ in $M$ centered at $x\in X$. 
For any $s\geq 0$, one can compute topological invariants of the space $X^{(s)}$ --- such as its homology and cohomology with different field coefficients \cite{hatcher2002algebraic} --- from finite combinatorial data encoded in the distances between points in $X$. 
One can then measure how the topological invariants change as the scale $s$ varies; if a topological feature is stable over a large range of scales, it is interpreted as reflecting a feature of the topological space hypothesized to underlie the sampling $X$. \\

\noindent The results of a persistent (co)homology computation are summarized in a so-called  persistence diagram $\mathsf{dgm} \subseteq [0,\infty]\times [0 ,\infty]$. 
Each point  in $\mathsf{dgm}$ represents a persistent (co)homology class of a particular dimension, and the $x$- and $y$-coordinates indicate the range of scales for which the corresponding class is alive.
Persistent classes of dimension-0 roughly correspond to clusters (connected components), dimension-1 classes correspond to 1-dimensional loops or holes, and 2-dimensional classes correspond to voids. 
   \\

\noindent \textbf{Sparse Circular Coordinates: }  Let $X$ be a subset of a metric space $M$. The sparse circular coordinates algorithm \cite{Sparse_CC} uses a 1-dimensional class from the persistent cohomology of  $X$ to produce a map $f:X^{(s)}\to\mathbb{S}^{1}$ which respects (represents) the persistent circular feature associated with that class. 
In addition to circular coordinates, there have been other theoretical developments, generalizations and applications to computer vision and signal processing~\cite{polanco2019lens}~\cite{scoccola2022toroidal},~\cite{scoccola2022fibered}. 
An implementation can be found in the DREiMac library for topological visualization, coordinatization and dimensionality reduction \cite{DREiMac}. \\

\noindent \textbf{Discrete Approximate Circle Bundles: }In topology, a \textit{fiber bundle} with fiber $F$ is a continuous surjection $p:E\to B$ between topological spaces, with the following property: for each $b\in B$, there is an open neighborhood $U_{b}\subseteq B$ of $b$ and a homeomorphism $\varphi:p^{-1}(U_{b}) \to  U_{b}\times F$ such that  if $p_b : U_b\times F \to U_b$ is the projection onto the first factor, then the following diagram commutes:

\begin{equation*}
\begin{tikzcd}
p^{-1}(U_{b}) \arrow[rr, "\varphi"] \arrow[dr, swap, "p"] & & U_{b}\times F \arrow[dl, "p_{b}"] \\
& U_{b} &
\end{tikzcd}
\end{equation*}

In a recent paper~\cite{turow2025discrete}, we introduced a discrete approximate version of a fiber bundle with fiber $\mathbb{S}^{1}$ (which we call a \textit{discrete approximate circle bundle}), and provide algorithms to classify and coordinatize these objects. 
In this paper, we apply the theory and algorithms found there to confirm the optical flow torus model proposed in~\cite{opt_flow_torus} as part of a larger 3-manifold structure.
See~\cite{turow2025discrete} for additional theoretical details about circle bundles and their discrete approximate analogs. \\

\section{Preprocessing}\label{sec: Preprocessing}

\noindent In order to obtain results comparable with \cite{Horizontal_Flow_Circle, opt_flow_torus}, we used   the same preprocessing steps, summarized below: \\

\noindent \textbf{Step 1:} Collect a random sample of 4,000 patches of size $3\times 3 $ from each frame of the Sintel video (for a total of $4.164\times 10^{6}$). 
Each $3 \times 3$ patch is originally a matrix of ordered pairs representing the horizontal and vertical components of the flow vectors at each pixel:

\begin{equation*}    
    \begin{pmatrix}
    (u_1, v_1) & (u_4, v_4) & (u_7, v_7) \\
    (u_2, v_2) & (u_5, v_5) & (u_8, v_8) \\
    (u_3, v_3) & (u_6, v_6) & (u_9, v_9)
    \end{pmatrix}
\end{equation*}

\noindent This matrix can then be flattened to a column vector $(u_{1},...,u_{9},v_{1},...,v_{9})^T \in\mathbb{R}^{18}$. \\

\noindent \textbf{Step 2:} Compute the \textbf{contrast norm} $||x||_{D}$ of each patch in the sample. Here contrast norm is defined as the sum of the squared differences between the components of flow vectors at (horizontally or vertically) adjacent pixels ($i\sim j$):

\begin{equation*}
    \| x\|_{D}^{2} = \sum_{i\sim j}(u_{i} - u_{j})^{2} + (v_{i} - v_{j})^2 
\end{equation*}

\noindent This notion of contrast for optical flow patches is inspired by a similar definition for range and optical image patches used in \cite{Lee_Mumford}, \cite{Klein_bottle}, and \cite{Range_Patches} (which is just the $u$ or $v$ portion of the formula above). 
One may view this distance as a discretization of the Dirichlet semi-norm

\begin{equation*}
    \|f\|_{D}^{2} = \iint\limits_{[-1, 1]^2} \|\nabla f(x,y)\|^{2} \, dx \, dy
\end{equation*}

\noindent which is the unique scale-invariant norm on images $f: [-1,1]^2 \to \mathbb{R}$. 
The combinatorics of the contrast norm are captured by a $9 \times  9$ symmetric matrix $D$ in the sense that $\|x\|_{D}^{2} = u^{T}Du\hspace{1mm} +\hspace{1mm} v^{T}Dv$ for all $x = (u,v)^T\in\mathbb{R}^{18}$ (see \cite{Lee_Mumford} for details).  \\

\noindent \textbf{Step 3:} Keep only those patches in the top 20 percent by contrast norm. 
Normalize said patches to have contrast norm 1 and mean flow 0. Note that one can always recover a patch from its normalized version given its  mean flow and contrast norm.  \\

\noindent \textbf{Step 4:} Downsample to $2.5\times 10^{5}$ patches for computational feasability and denote this sample by $X$. Note that $X$ is five times larger than the samples used in \cite{Horizontal_Flow_Circle}, \cite{opt_flow_torus}.\\

\noindent \textbf{Step 5:} Restrict to a dense core subset. Specifically, for $x\in X$, let $\rho_{k}(x)$ denote the distance (in $\mathbb{R}^{18}$) from $x$ to its $k^{\text{th}}$ nearest neighbor in the sample. Let $X(k,q)$ denote the top $q$ percent of patches in $X$ by density, using $1/\rho_{k}(x)$ as a density estimator. We will mostly consider the dense core subsets $X(1500,50)$ and $X(50,60)$ (which roughly correspond to $X(300,50)$ and $X(10,60)$ for the smaller samples used in \cite{Horizontal_Flow_Circle} and \cite{opt_flow_torus}). \\   

\noindent Note that the space of mean-centered, contrast normalized patches in $\mathbb{R}^{18}$ is an ellipsoid $S$ homeomorphic to $\mathbb{S}^{15}$. We therefore have no reason to expect a priori that a significant portion of $X$ is concentrated around a 2- or 3-dimensional manifold.  

\vspace{5mm}
\section{The Torus Model}\label{sec: Torus Model}

In this section, we review aspects of the torus model and results from \cite{opt_flow_torus} which are directly relevant to our subsequent investigation and findings. We point out areas in which the model seems to disagree with experiment, providing motivation for our extended model. \\

\noindent In \cite{Lee_Mumford}, the authors used a DCT (Discrete Cosine Transform) basis to model high-contrast optical image and range patches:

\begin{figure}[h!]
  \centering
  \includegraphics[width=0.75\linewidth]{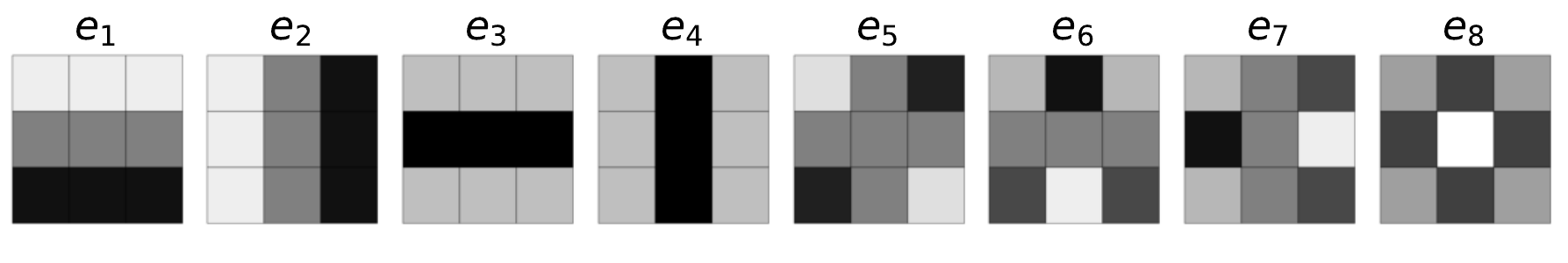}
  \caption{The mean-centered contrast-normalized DCT basis for $3 \times 3$ image patches}   
  \label{Range_DCT}
\end{figure}

\noindent These patches form an eigenbasis for the $D$ matrix used to compute the contrast norm. 
Note that the constant patch $(1,...,1)^{T}\in\mathbb{R}^{9}$ is also an eigenvector of $D$, though it has contrast norm 0. 
By definition, mean-centered data lies in the span of the other eigenvectors. \\

\noindent In terms of this basis, the primary circle model for $3 \times 3$ high contrast range and optical image patches described in \cite{Klein_bottle} is precisely the circle spanned by $e_{1}$ and $e_{2}$ -- see Figure~\ref{Range_Step_Annulus}.\\

\noindent The corresponding eigenvectors of the block diagonal matrix $\begin{bmatrix}D & 0 \\ 0 & D \end{bmatrix}$ form a natural basis for mean-centered optical flow patches: \\

\begin{figure}[h!]
  \centering
  \includegraphics[width=0.7\linewidth]{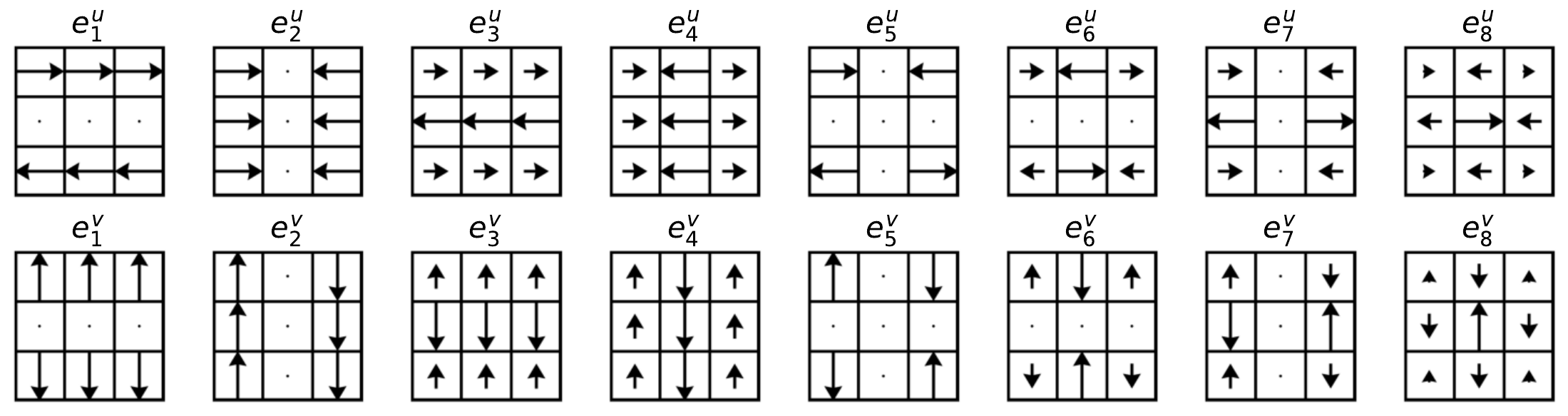}
  \caption{Optical flow DCT basis. }   
  \label{Flow_DCT}
\end{figure}

\noindent Using a sample size of $5\times 10^{4}$ patches, \cite{Horizontal_Flow_Circle} provides evidence that $X(300,30)$ is dense around the \textit{horizontal flow circle} obtained by applying horizontal camera motion to patches in the primary circle.
In terms of the optical flow DCT basis, the horizontal flow circle is precisely the circle spanned by $e_{1}^{u}$ and $e_{2}^{u}$. 
The authors hypothesize that high-contrast optical flow patches are the result when camera translations are applied to boundaries between foreground and background in a scene (i.e., high-contrast range patches), and that horizontal translations are simply the most prevalent. \\

\begin{figure}[h!]
  \centering
  \includegraphics[width=\linewidth]{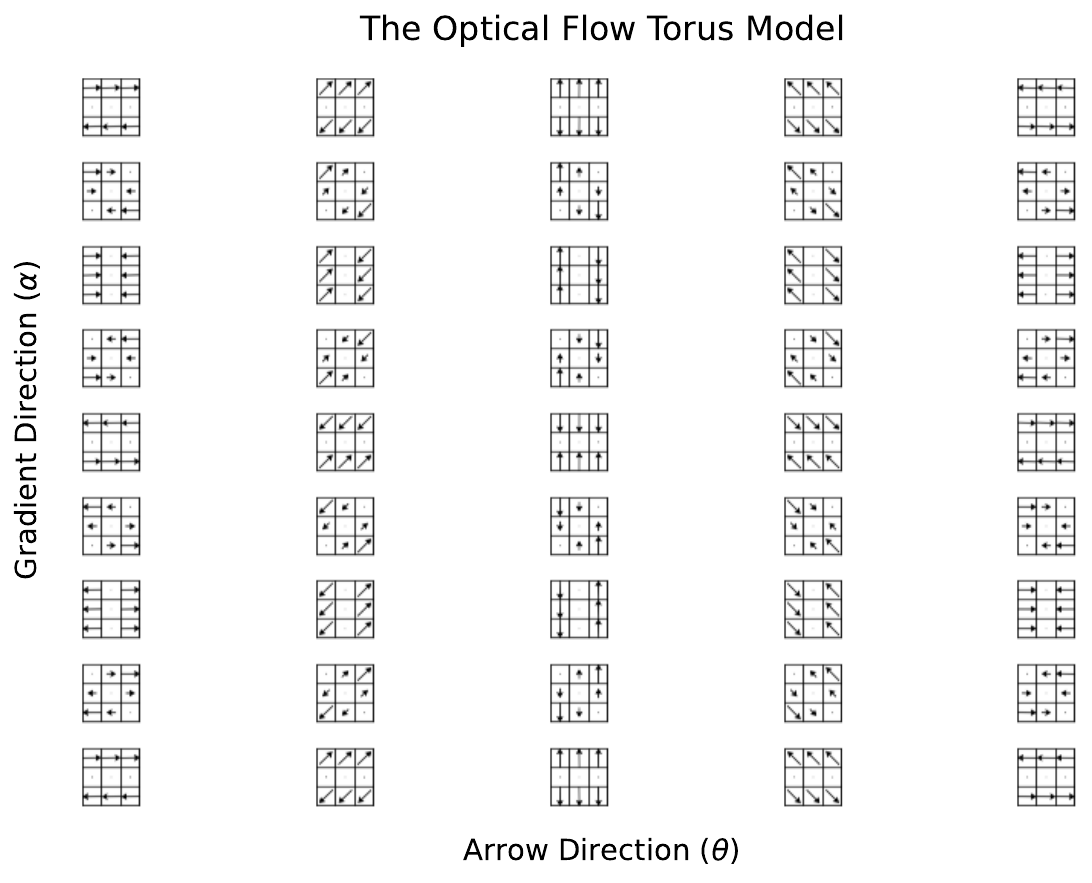}
  \caption{The proposed optical flow torus model. Each column captures patches with the same flow direction. The horizontal flow circle is shown in the first and last column; note however that the two columns differ by a shift by $\pi$. This reflects the fact that $(\theta,\alpha)$ is not a global coordinate system for $\mathcal{T}$. Compare with Figure 7 in~\cite{opt_flow_torus}.} 
  \label{Optical_Flow_Torus}
\end{figure}

\noindent In \cite{opt_flow_torus}, the authors assert that $X(300,30)$ actually contains a  direction flow circle for each 1-dimensional subspace of $\mathbb{R}^{2}$, and that these circles stitch together to form a torus. 
Intuitively, each circle corresponds to camera translation along a particular axis (see Figure~\ref{Optical_Flow_Torus}). 
The model they propose is a submanifold of the 3-sphere spanned by $e_{1}^{u}$, $e_{2}^{u}$, $e_{1}^{v}$ and $e_{2}^{v}$, and is described by the image of the map $F:\mathbb{T}^{2}\to\mathbb{R}^{18}$ defined by

\begin{equation}\label{eq: torus model}
    F(\alpha, \theta) = \cos{(\theta)}\left(\text{cos}{(\alpha)}e_{1}^{u} + \text{sin}{(\alpha)}e_{2}^{u}\right) + \sin{(\theta)}\left(\text{cos}{(\alpha)}e_{1}^{u} + \text{sin}{(\alpha)}e_{2}^{u}\right)
\end{equation}

\vspace{3mm}

\noindent We denote the image of $F$ by $\mathcal{T}$. Note that $F$ is actually a double cover of $\mathcal{T}$ with $F(\alpha + \pi, \theta + \pi) = F(\alpha, \theta)$. We obtain an embedding of $\mathbb{T}^{2}$ into $\mathbb{R}^{18}$ by defining $G(\omega, \theta) = F(\omega - \theta, \theta)$ (see Figure~\ref{Recovered_Patch_Diagrams}).  \\

\newpage
\subsection{Evidence For The Torus Model}\label{sec: Evidence For Torus}

\begin{figure}[h!]
  \centering
  \includegraphics[width=0.8\linewidth]{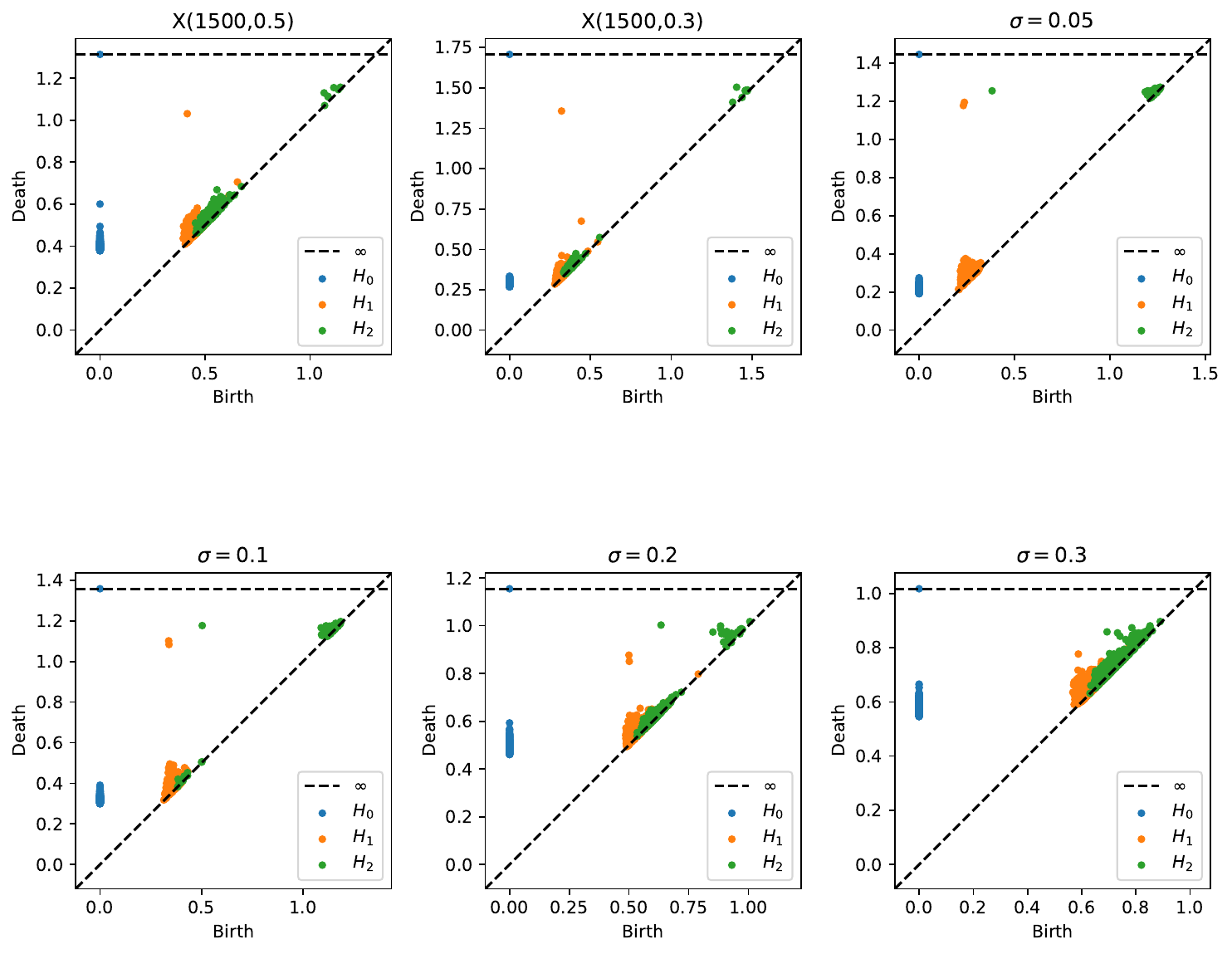}
  \caption{Persistence diagrams for $X(1500,50)$, $X(1500,30)$ and several samples from the idealized torus model $\mathcal{T}$ with various levels of Gaussian noise}   
  \label{Global_Ripsers}
\end{figure}

A direct  persistent homology computation for $X(1500, 30)$ or $X(1500, 50)$ does not provide strong evidence for the existence of a torus.
Indeed,
Figure~\ref{Global_Ripsers} shows the persistence diagrams for random samples of 500 points from $X(1500,30)$ and $X(1500,50)$ as well as several synthetic samples from the model torus with various levels $\sigma$ of Gaussian noise (we found the results to be robust under this downsampling). Observe that there is no significant range of scales for which either of the real samples has the Betti signature of a torus (namely $\beta_{0} = 1, \beta_{1} = 2, \beta_{2} = 1$). In particular, the persistence diagrams for the real samples do not resemble the diagrams for any of the noisy synthetic samples (i.e., from $\mathcal{T}$). 
\vspace{2mm}

\noindent To provide evidence for the torus model, Adams et al. introduce a feature map $p$ which assigns a 1-dimensional linear subspace of $\mathbb{R}^2$, $p (x)\in\mathbb{RP}^{1} $ (called the \textit{predominant direction}), to each patch $x$ in a neighborhood of the model torus $\mathcal{T}$. 
Loosely speaking, $p$ identifies the linear subspace of $\mathbb{R}^{2}$ which is most colinear to the arrows of a flow patch. 
In particular, the restriction of $p$ to $\mathcal{T}$ is a fiber bundle with fiber $\mathbb{S}^{1}$ (see Methods). 

\begin{figure}[h!]
    \centering
    \includegraphics[width=\textwidth]{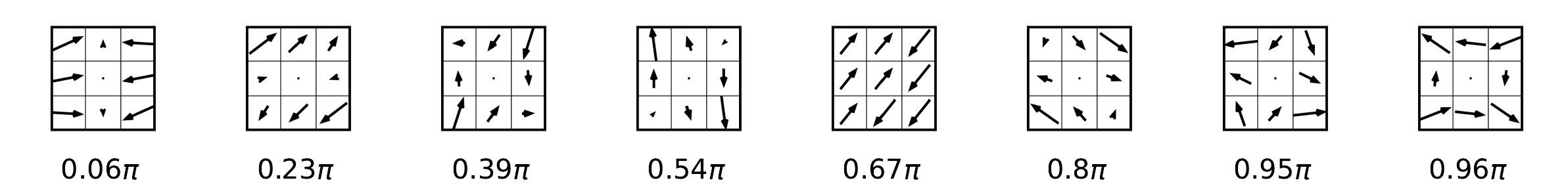}
    \caption{A sample of high-contrast optical flow patches labeled by predominant flow direction}
    \label{sample_predom_dirs}
\end{figure}

\noindent Following \cite{opt_flow_torus}, we divided $X(1500,50)$ into six  subsets of the form $X_{\theta}(1500,50) = p|_{X(1500,50)}^{-1}\left(B_{\frac{\pi}{12}}(\theta)\right)$ for $\theta = \frac{k\pi}{6},\hspace{2mm}k = 1, ..., 6$, then computed the orthogonal projection of each set onto the plane spanned by the vectors $\left(\text{cos}(\theta)e_{1}^{u} + \text{sin}(\theta)e_{1}^{v}\right)$ and $\left(\text{cos}(\theta)e_{2}^{u} + \text{sin}(\theta)e_{2}^{v}\right)$ (for the appropriate value of $\theta$):\\

\begin{figure}[h!]
    \centering
    \includegraphics[width=0.75\textwidth]{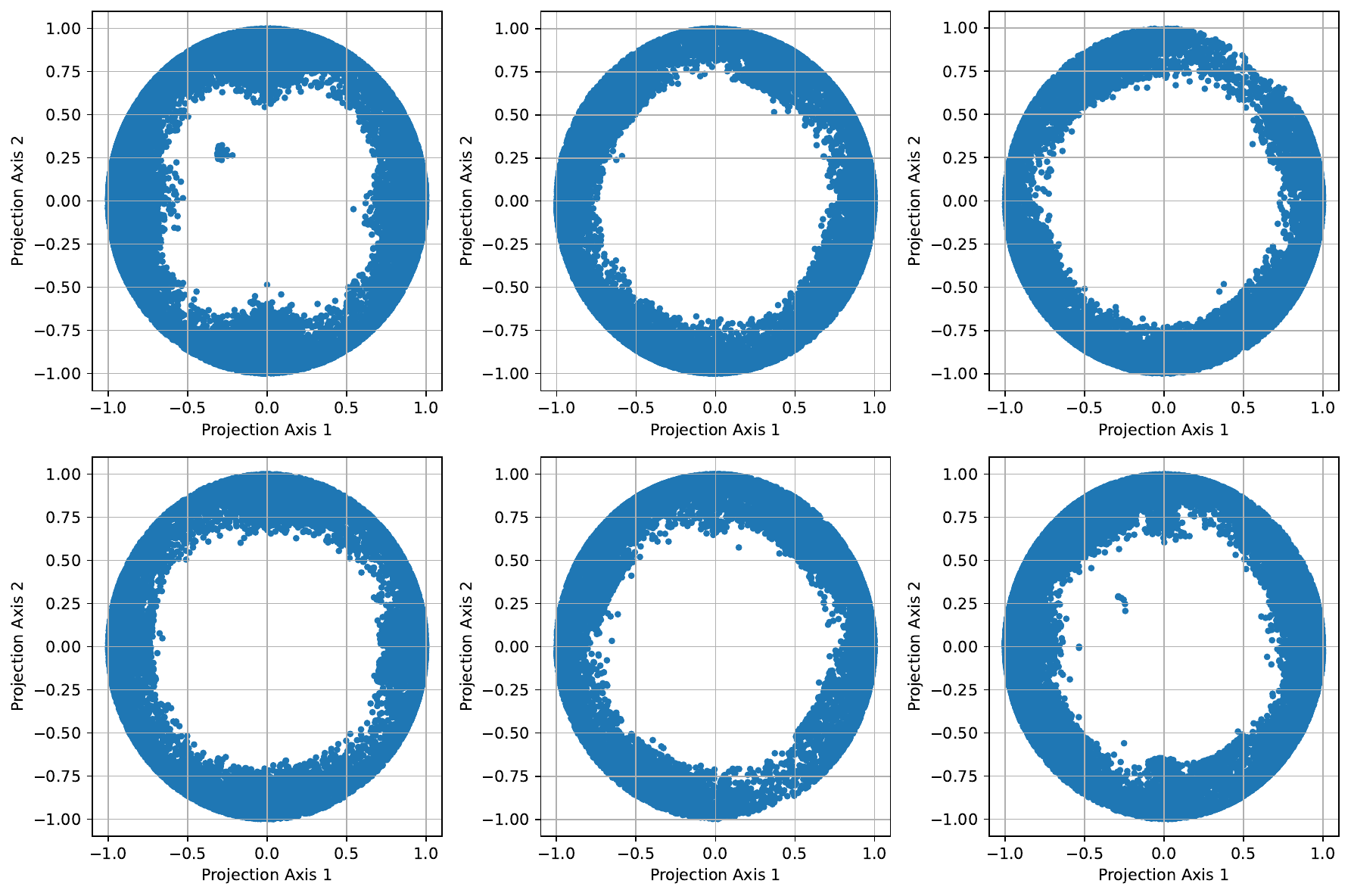}
    \caption{Plane projections of the sets $\{X_{\frac{k\pi}{6}}(1500,50)\}_{k=1}^{6}$}
    \label{Adams_fiber_projections}
\end{figure}

\noindent Keeping in mind that the data has been contrast-normalized to lie on the ellipsoid $S\cong \mathbb{S}^{15}$ spanned by the optical flow DCT basis, the projections above strongly suggest that the data in each $X_{\theta}(1500,50)$ is concentrated near the circle $p^{-1}(\theta)$. 
The authors of \cite{opt_flow_torus} also provide a zigzag persistent homology computation \cite{Zigzag} as evidence that the circular features in the different fibers of $p$ stitch together compatibly to form a fiber bundle structure. 
However, as the authors note, this computation is insufficient to distinguish between a torus and a Klein bottle (the two possible global topologies for the total space of a fiber bundle over $\mathbb{RP}^{1}$ with fiber $\mathbb{S}^{1}$) because it does not designate or track relative orientations for generators of local homology. 

\vspace{2mm}

\subsection{The Predominant Direction Map $p$}\label{sec: Predom Dir Map}

\noindent Given a point $x\in S\subset\mathbb{R}^{18}$, the \textbf{predominant direction} of $x$ is formally defined in \cite{opt_flow_torus} to be the line $p(x)\in\mathbb{RP}^{1}$ spanned by the first singular vector of the $9\times 2$ matrix $A_{x}$ whose rows are the component flow vectors $\{(x_{i}, x_{i+9})\}_{i=1}^{9}$ which comprise $x$. 
In particular, the assignment $x\mapsto p (x)$ is continuous and surjective on a neighborhood of $\mathcal{T}$ and satisfies $p (F(\theta, \alpha)) = \theta \ (\text{mod } \pi)$ for all $(\theta,\alpha)\in\mathbb{S}^{1}\times\mathbb{S}^{1}$ (where $F$ is the double cover of $\mathcal{T}$ shown in  Equation~\eqref{eq: torus model}).\\

\noindent Note that $p(x)$ is not well-defined if the singular values of $A_{x}$ are equal, which occurs when the component arrows of $x$   point in  different directions. 
If  the data is concentrated around the torus model, as theorized in \cite{opt_flow_torus},  predominant direction is a meaningful measurement to assign to any patch $x\in X(1500,50)$.
However, this does not hold for a large portion of the data. 
Indeed, there are many regions in $S$ for which  predominant direction is undefined. 
Even if the singular values associated to a flow patch are only approximately equal, the assignment of a predominant direction does not have a clear meaning:

\begin{figure}[ht]
    \centering
    \includegraphics[width=0.49\textwidth]{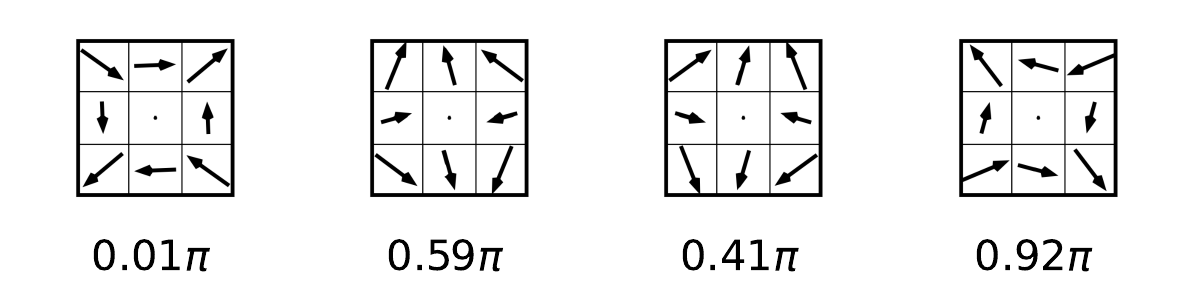}
    \hspace{1mm}
    \includegraphics[width=0.49\textwidth]{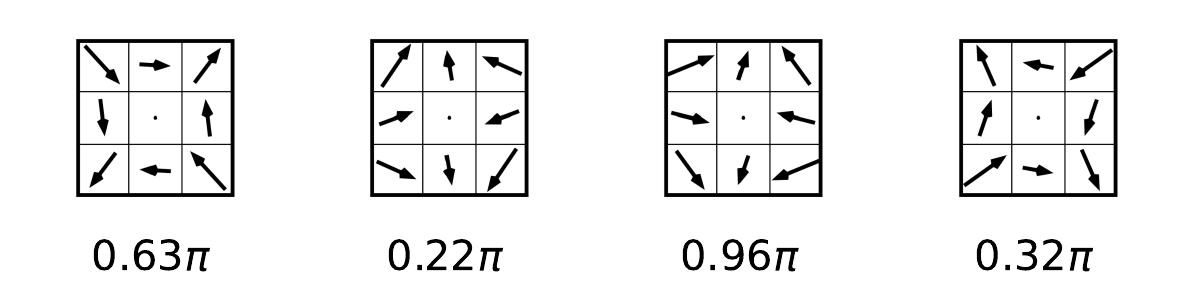}
    \caption{A sample of high-contrast optical flow patches from $X(1500,50)$ for which the predominant direction is essentially arbitrary} 
    \label{Weak_Predom_Dir_Samples}
\end{figure}

\noindent To assess the extent to which predominant direction is well-defined on patches in $X(1500,50)$, we define the \textbf{directionality} of a patch $x\in S$ by the formula below: 
\vspace{2mm}

\begin{equation}\label{eq: Directionality}
    r(x) = \frac{|\lambda_1 - \lambda_2|
    }{\text{max}(\lambda_{1},\lambda_{2})}
\end{equation}

\vspace{2mm}
\noindent where $\lambda_{1}$ and $\lambda_{2}$ are the eigenvalues of $A_{x}^{T}A_{x}$ (equivalently, the squares of the singular values of $A_{x}$). Note that $r(x) = 0$ iff the eigenvalues are equal and $r(x)$ approaches 1 as the difference between the eigenvalues increases (or if one of the eigenvalues is 0). In particular, $r(x) = 1$ for all patches in $\mathcal{T}$. Since $0\notin S$, $r$ is well-defined and continuous for all $x\in S$. \\

\begin{figure}[h!]
    \centering
    \includegraphics[width=\textwidth]{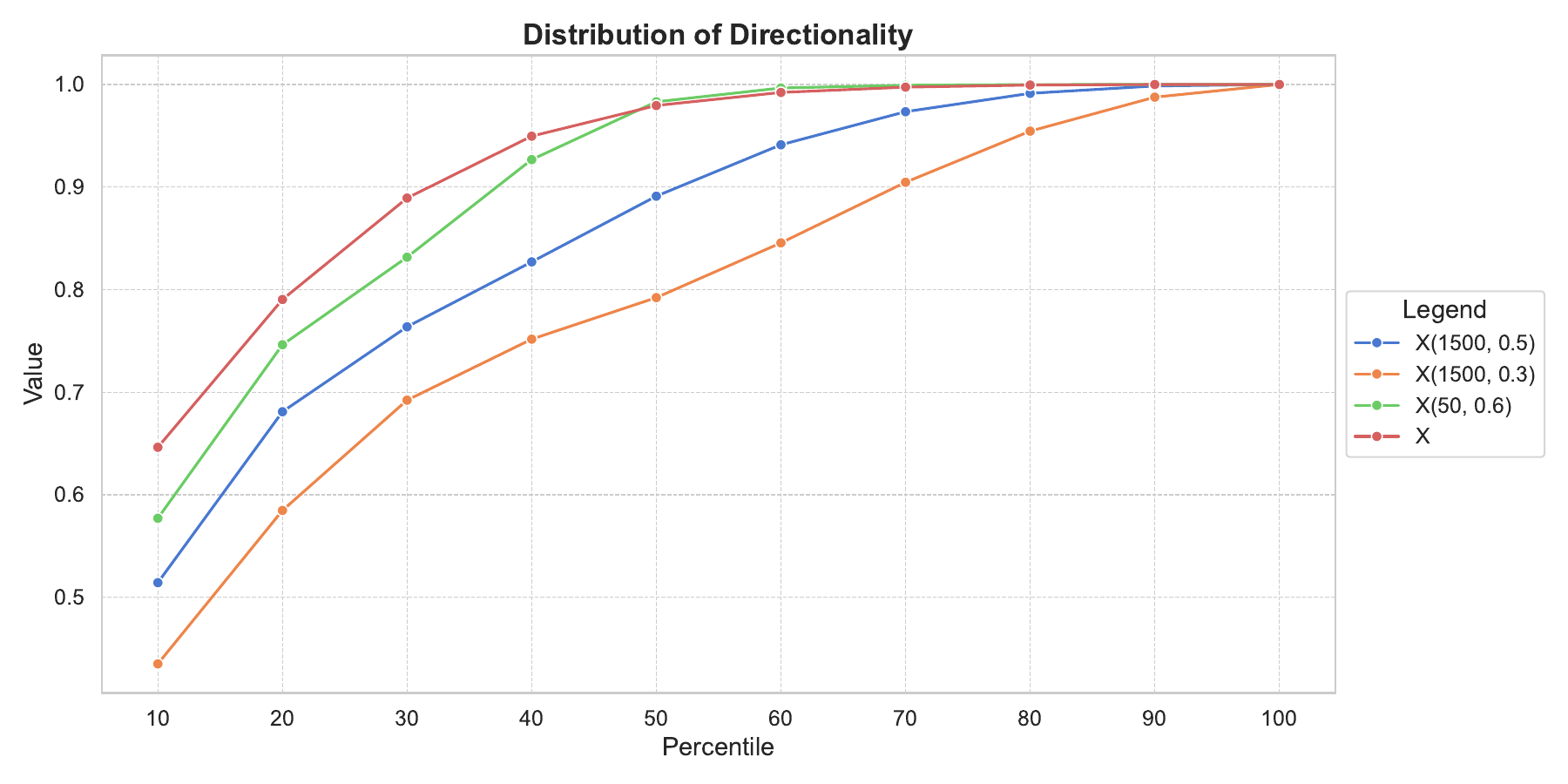}
    \caption{ }
    \label{Distributions_Of_Directionality}
\end{figure}

\noindent Figure~\ref{Distributions_Of_Directionality} shows the distribution of directionality across the dataset $X$ and various core subsets. Observe that each distribution is skewed towards 1, though a substantial portion of the data has notably lower directionality. \\

\noindent Note in particular that the percentile graph of $r$ for $X(1500,30)$ is strictly below the corresponding graph for $X(1500,50)$. We conclude that the patches of lower directionality are an integral part of the dense core subset and cannot be ignored.  

\vspace{5mm}
\section{Extended Model}\label{sec: Extended Model}

In this section, we propose an extended model for the dense core subset $X(1500,50)$ which includes patches of low-directionality. We show that our improved model is more topologically and geometrically faithful to the true structure of the data, and that it explains why the torus model could not be verified with a direct persistence computation (Figure \ref{Global_Ripsers}). \\ 

\noindent Recall that the optical flow torus was defined as the image of a map $F:\mathbb{S}^{1}\times\mathbb{S}^{1}\to \mathcal{T}\subset\mathbb{R}^{18}$ (see Equation~\eqref{eq: torus model}). Our proposed extended model for $X(1500,50)$, which includes patches of lower directionality, is the image of the map $\tilde{F}:(\mathbb{D}^{2}/\{0\})\times\mathbb{S}^{1}\to\tilde{\mathcal{T}}\subset \mathbb{R}^{18}$ defined by

\vspace{3mm}

\begin{equation}
    \tilde{F}(r,\alpha,\theta) = \cos{\left(\tau(r)\right)} F(\alpha, \theta) +  \sin{\left(\tau(r)\right)} F^{\perp}(\alpha, \theta)
\end{equation}

where 

$$F^{\perp}(\alpha,\theta) = F\left(\alpha + \frac{\pi}{2}, \theta - \frac{\pi}{2}\right),\hspace{2cm} \tau(r) = \text{cos}^{-1}\left((2 - r)^{-\frac{1}{2}}\right)$$

\vspace{3mm}
for all $r\in (0,1]$ and $\theta,\alpha\in\mathbb{S}^{1}$.  

\vspace{5mm}
\textbf{Proposition 1:} The map $\tilde{F}$ has the following key properties:

\vspace{3mm}
\begin{enumerate}
    \item For all $r,\alpha,\theta$, the patch $\tilde{F}(r,\alpha,\theta)$ has mean 0, contrast norm 1, predominant direction $\theta$ ($\text{mod }\pi$) and directionality $r$.
    \vspace{3mm}

    \item $\tilde{F}(1,\alpha,\theta) = F(\alpha,\theta)$ for all $\alpha,\theta$, so $\tilde{F}|_{\mathbb{T}^{2}} = F$.
    \vspace{3mm}

    \item $\tilde{F}$ is a double cover of its image $\tilde{\mathcal{T}}$ with $\tilde{F}(r,\alpha + \pi, \theta + \pi) = \tilde{F}(r,\alpha, \theta)$. Thus, the map $\tilde{G}:\mathbb{RP}^{1}\times (0,1]\times \mathbb{S}^{1}\to \tilde{\mathcal{T}}$ defined by $\tilde{G}(r,\omega,\theta) = \tilde{F}(r,\omega - \theta, \theta)$ is a homeomorphism.
    \vspace{3mm}

    \item $F^{\perp}(\alpha,\theta)$ is the inward-pointing unit normal vector to $\mathcal{T}$ (as a hypersurface of the 3-sphere spanned by $\{e_{1}^{u},e_{2}^{u},e_{1}^{v},e_{2}^{v}\}$) at $F(\alpha, \theta)$.   
    \vspace{3mm}
    
    \item For all $\alpha,\theta$, we have 

\begin{equation}
    \lim_{r\to 0}\tilde{F}(r,\alpha,\theta) = \text{cos}(\alpha + \theta)\left(\frac{e_{1}^{u}-e_{2}^{v}}{\sqrt{2}}\right) + \text{sin}(\alpha + \theta)\left(\frac{e_{2}^{u}+e_{1}^{v}}{\sqrt{2}}\right)    
\end{equation}

\end{enumerate}

\begin{proof}
  See Appendix~\ref{sec: Proof Of Proposition}.
\end{proof}

\noindent Intuitively, our model characterizes the ground truth manifold for high-contrast optical flow patches as a solid torus with the central circle removed, whose boundary is $\mathcal{T}$. In particular, directionality manifests geometrically as a radial degree of freedom perpendicular to $\mathcal{T}$ which is collapsed by the predominant direction map $p$. Thus, the fibers of the fiber bundle $p:\tilde{\mathcal{T}}\to\mathbb{RP}^{1}$ are homeomorphic to cylinders rather than circles, and from the perspective of persistent homology, which relies on the ambient metric, the total space resembles a circle rather than a torus.  

\begin{figure}[h!]
  \centering
  \includegraphics[width=0.45\linewidth]{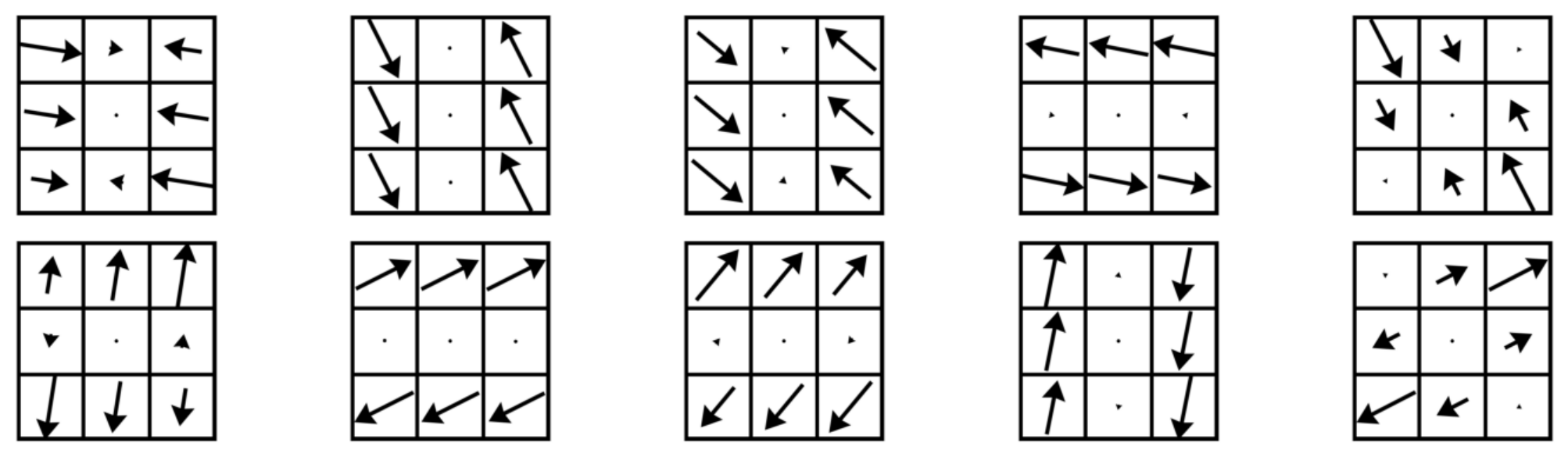}
  \caption{$F(\alpha,\theta)$ (top) and $F^{\perp}(\alpha,\theta)$ (bottom) for various $(\alpha,\theta)$ pairs.} 
  \label{Antipatches}
\end{figure}

\noindent Note that $F^{\perp}(\alpha, \theta)$ is the patch in $\mathcal{T}$ obtained from $F(\alpha,\theta)$ by rotating the predominant direction by $\frac{\pi}{2}$ and rotating the gradient direction by $\frac{\pi}{2}$ in the opposite direction (as long as the rotations are in opposing directions, the final result will be the same). Patches of lower directionality are obtained by mixing an optical flow torus patch $F(\alpha,\theta)$ with its perpendicular counterpart $F^{\perp}(\alpha,\theta)$ in varying proportions. If $F(\alpha,\theta)$ and $F^{\perp}(\alpha,\theta)$ are mixed in equal proportions, we obtain a patch with zero directionality:  \\  

\begin{figure}[ht]
    \centering
    \begin{subfigure}{0.48\textwidth}
        \centering
        \includegraphics[width=\linewidth]{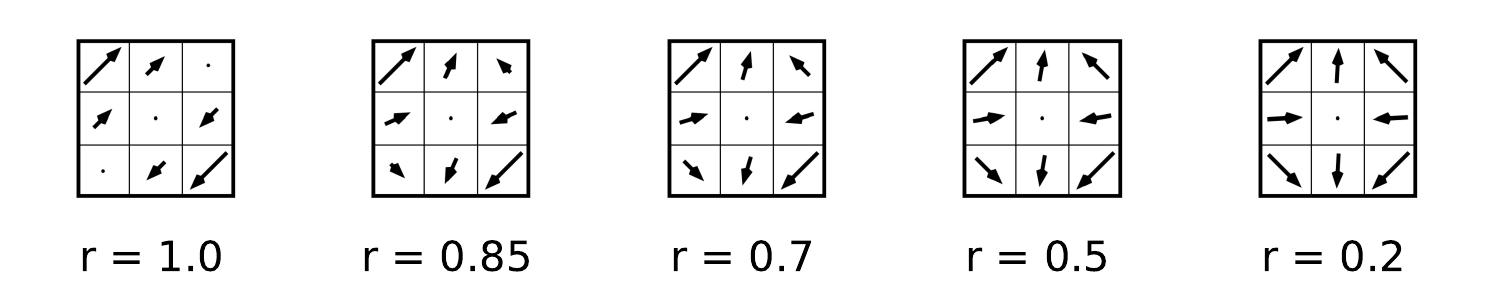}
        \caption{$\alpha_{0}=\theta_{0}=\frac{\pi}{4}$}
    \end{subfigure}
    \hspace{2mm}
    \begin{subfigure}{0.48\textwidth}
        \centering
        \includegraphics[width=\linewidth]{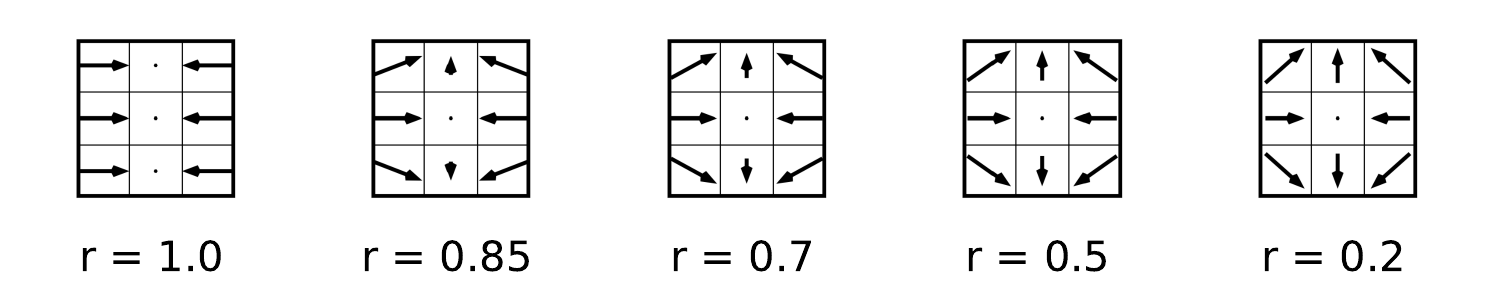}
        \caption{$\alpha_{0} = \frac{\pi}{2}, \ \theta_{0} = 0$}
    \end{subfigure}
    \caption{Model flow patches of the form $\widetilde{F}(r,\alpha_{0},\theta_{0})$ for decreasing values of $r$. Note that $\widetilde{F}(1,\alpha_{0},\theta_{0}) = F(\alpha_{0},\theta_{0})$ is a flow torus patch, and the limiting patch as $r\to 0$ depends only on the sum of $\alpha_{0}$ and $\theta_{0}$.} 
    \label{Decreasing_Directionality}
\end{figure}

\noindent Property 5 of Proposition 1 implies that patches with low-directionality are concentrated around a single circle. In fact, as one restricts to patches of lower and lower directionality, each fiber of $p$ collapses to this circle:  

\begin{figure}[h!]
  \centering
  \includegraphics[width=0.3\linewidth]{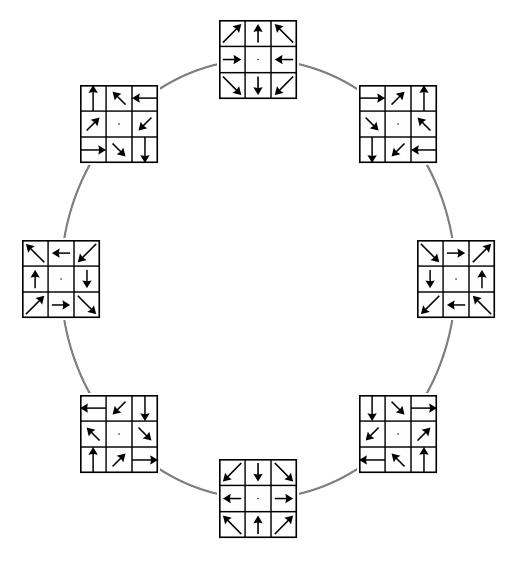}
  \caption{The limiting circle of patches with low directionality in the proposed extended model}

  \label{Circle_Of_Low_Directionality}
\end{figure}

\subsection{Evidence For The Extended Model}\label{sec: Extended Model Evidence}

\noindent To verify our model, we used the pipeline proposed in~\cite{turow2025discrete} to certify the global triviality of the bundle structure underlying the data in $X(1500,50)$ and construct a global parametrization. Our adaptation of the procedure to the present context is summarized below: \\

\begin{enumerate}
    \item \textbf{Establish An Open Cover:} Choose $N$ equally-spaced landmarks \[\left\{l_{j} = \left[e^{\frac{\pi i}{N}j}\right]\right\}_{j=1}^{N}\subset\mathbb{RP}^{1} = \mathbb{S}^1/z \sim -z\]  and cover $\mathbb{RP}^{1}$ with the open  balls  $\mathcal{U} = \{U_{j} = B_{r}(l_{j})\}_{j=1}^{N}$, (i.e., circular arcs of length $\arcsin(r)$) where $r = \sin\left(\frac{\pi}{2N}(1+c)\right)$ for some constant $0 < c \leq \frac{1}{2}$. 
    Let 
    \[
    \mathcal{N}(\mathcal{U})
    =
    \left\{ \sigma \subseteq \{1,\dots, N\} \; : \; \bigcap_{j \in \sigma} U_j \neq \emptyset
    \right\}
    \]
    be the nerve of the open cover $\mathcal{U}$ -- note that $c\leq \frac{1}{2}$ implies $\mathcal{N}(\mathcal{U})$ contains no 2-simplices. 
    \vspace{5mm}
    
    \item \textbf{Construct Discrete Approximate Local Trivializations:} Compute the 1-dimensional persistent cohomology of $X_{j} = X\cap p^{-1}(U_{j})$, $j=1,\ldots,N$, and use the sparse circular coordinates algorithm from \cite{Sparse_CC} to construct maps $\{f_{j}:X_j\to\mathbb{S}^{1}\}_{j=1}^{N}$ which capture the most prominent 1-dimensional persistent cohomology class of each $X_j$.
    \vspace{5mm}
    
    \item \textbf{Compute Discrete Approximate Transition Maps:} For each $(jk) \in \mathcal{N(U)}$, use Procrustes alignment~\cite{schonemann1966generalized} to compute a matrix $ \Omega_{jk}\in O(2) $ minimizing $\sum\limits_{x\in X_j \cap X_k}\left|f_{j}(x) -  \Omega_{jk}f_{k}(x)\right|^2$.
    The family $\Omega = \{\Omega_{jk}\}_{(jk)\in\mathcal{N}(\mathcal{U})}$ can be interpreted as a locally-constant \v{C}ech $O(2)$-cocycle $\Omega\in \check{Z}^{1}\left(\mathcal{U};\underline{O(2)}\right)$ which represents a circle bundle over $\mathbb{RP}^{1}$ (every cochain is a cocycle since $\mathcal{N}(\mathcal{U})$ contains no 2-simplices). 
    See \cite{turow2025discrete} for the relevant definitions.
    \vspace{5mm}
    
    \item \textbf{Check For Orientability:} Compute the orientation (i.e., the Stiefel-Whitney) class representative $\omega   =  \{\omega_{jk} = \det(\Omega_{jk})\}_{(jk) \in \mathcal{N(U)}}  \in Z^1(\mathcal{N(U)}; \mathbb{Z}_2)$, where $\mathbb{Z}_2 = \{-1, 1\}$ is regarded as a multiplicative group. 
    By Proposition 2.13 in~\cite{turow2025discrete}, the circle bundle over $\mathbb{RP}^{1}\cong\mathbb{S}^{1}$ represented by $\Omega$ is isomorphic to either the torus or the Klein bottle, and $\Omega$ represents a torus if and only if $\omega$ is a simplicial coboundary. 
    That is, if there is a set  $\tau = \{\tau_j\}_{j=1}^N$,  $\tau_j \in \mathbb{Z}_2$,   so that $\omega_{jk} = \tau_j\tau_k$ for all $(jk) \in \mathcal{N(U)}$. 
    If $\tau$ exists, it is called a potential for $\omega$. 
    \vspace{5mm}

    \item \textbf{Synchronize Local Coordinates:} If $\omega$ is a coboundary (as we expect), then $\Omega$ must also be an approximate coboundary, so we can align our local coordinate systems to create a global toroidal coordinate system. 
    Indeed, we use Singer's method \cite{Singer} to construct a discrete approximate potential 
    $\mu = \left\{\mu_j \in O(2)\right\}_{j=1}^N$
    for $\Omega$ --- i.e., so that $\Omega_{jk} \approx \mu_j^T \mu_k$ --- and choose a partition of unity $\{\phi_{j}:U_{j}\to\mathbb{R}\}_{j=1}^{N}$ subordinate to $\mathcal{U}$. 
    For $1\leq j\leq N$ and $x\in X_{j}$, let $\tilde{f}_{j}(x) \in \mathbb{S}^1$ be the weighted Karcher mean \cite{karcher1977riemannian} of $\{\mu_{j}f_{k}(x)\}_{k=1}^{N}\subset\mathbb{S}^{1}$ with weights $\{\phi_{k}\circ p(x)\}_{k=1}^{N}$ (note that a weight of $0$ is assigned to $\mu_{k}f_{k}(x)$ if $x\notin X_{k}$). Since the $\tilde{f}_{j}$'s agree on the intersections $X_j \cap X_k$, they combine to form a global trivialization map $F:X\to \mathbb{R}\mathbb{P}^{1}\times\mathbb{S}^{1}$ defined by $F(x) = (p(x), \tilde{f}_{j}(x))$ whenever $x\in X_{j}$ for some $j$.\\
\end{enumerate}

\noindent For details about the mathematical theory underlying the procedure described above, see our companion paper about discrete approximate circle bundles~\cite{turow2025discrete}.\\

\subsection{Results}\label{sec: Results}
\vspace{2mm}
For our open cover, we chose $N = 16$ to ensure each $U_{j}$ would be sufficiently small so that $X_{j} = X\cap p^{-1}(U_{j})$ could be treated as a 'fiber', but large enough that each such set would be sufficiently populated for analysis. Similarly, we chose $c = \frac{1}{2}$ to ensure that double intersections between open sets would be as large as possible without creating triple intersections. \\ 

\begin{figure}[h!]
    \centering
    \begin{subfigure}{.4\textwidth}
        \centering
        \includegraphics[width=\linewidth]{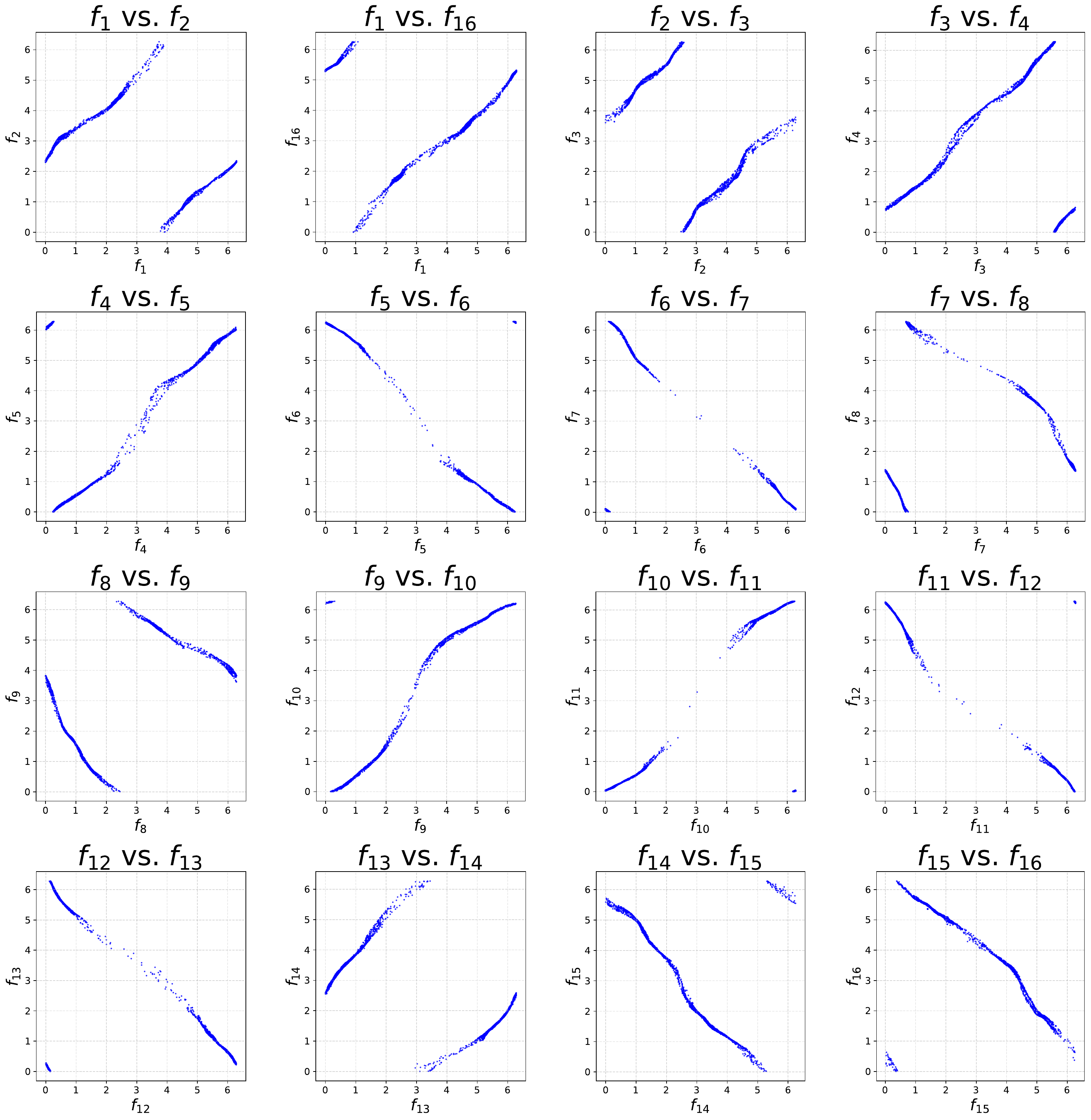}
    \end{subfigure}
    \hspace{0.05\textwidth}
    \begin{subfigure}{.4\textwidth}
        \centering
        \includegraphics[width=\textwidth]{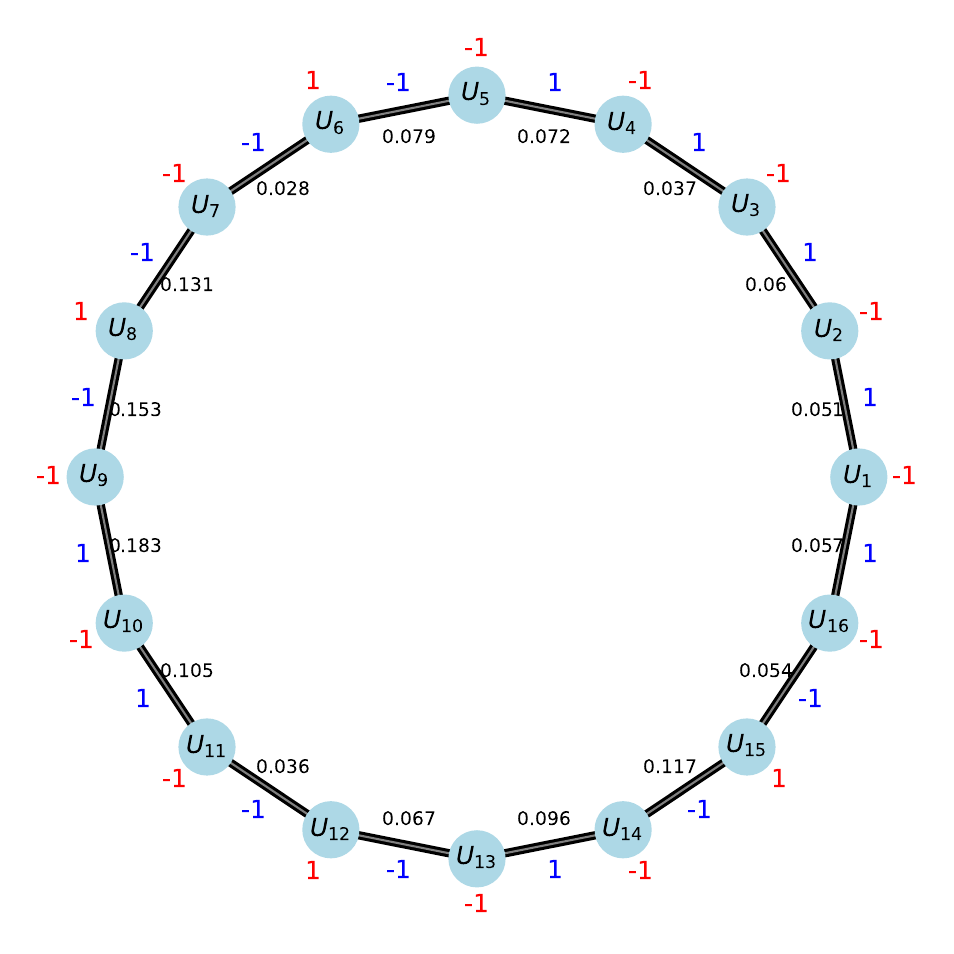}
    \end{subfigure}
    \caption{(Left) Correlations on intersections $X_j\cap X_k$ between the discrete approximate local trivializations $\{f_{j}:X_j \to \mathbb{S}^{1}\}_{j=1}^{16}$ computed for the data. (Right) The nerve of the open cover $\mathcal{U} = \{U_{j}\}_{j=1}^{16}$ of $\mathbb{RP}^{1}$. 
    Each edge $(jk)\in \mathcal{N(U)}$ is labeled by the average angular alignment error $\frac{1}{\#(X_j \cap X_k)}\sum\limits_{x\in X_j \cap X_k} |f_j(x) - \Omega_{jk}f_k(x)|^2$ of the local trivializations on the corresponding intersection. 
    The components $\omega_{jk}$ of the Stiefel-Whitney class representative $\omega =\text{det}_{*}(\Omega)$ are shown in blue, and a potential $\tau = \{\tau_{j}\}_{j=1}^N$ for $\omega$ is shown in red.}
    \label{Nerve_With_Cochains}
\end{figure}

\noindent Figure~\ref{Nerve_With_Cochains} shows a visualization of the nerve $\mathcal{N}(\mathcal{U})$ of the open cover $\mathcal{U}$. As expected, we found that $\omega = \text{det}_{*}(\Omega)$ was a coboundary. 

\begin{figure}[h!]
  \centering
  \includegraphics[width=0.65\linewidth]{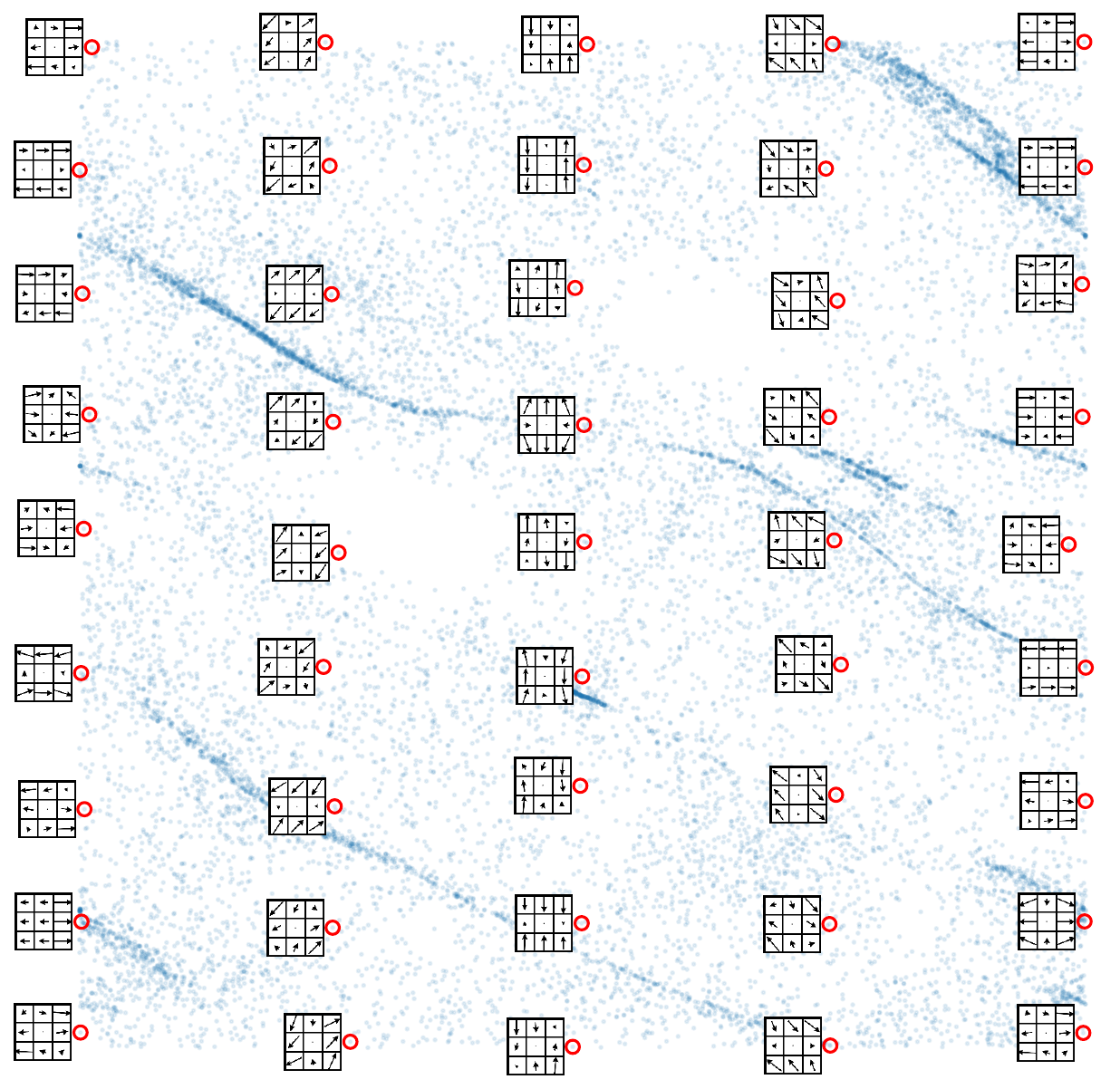}
  \caption{A sample of coordinatized flow patches from $X(1500,50)$ with high directionality, arranged according to predominant direction (x-axis) and assigned fiber coordinate (y-axis). Observe that the assigned fiber coordinate is strongly correlated with $\theta - \alpha$ in Figure~\ref{Optical_Flow_Torus}.}
    \label{Recovered_Patch_Diagrams}
\end{figure}

\begin{figure}[h!]
  \centering
  \includegraphics[width=0.65\linewidth]{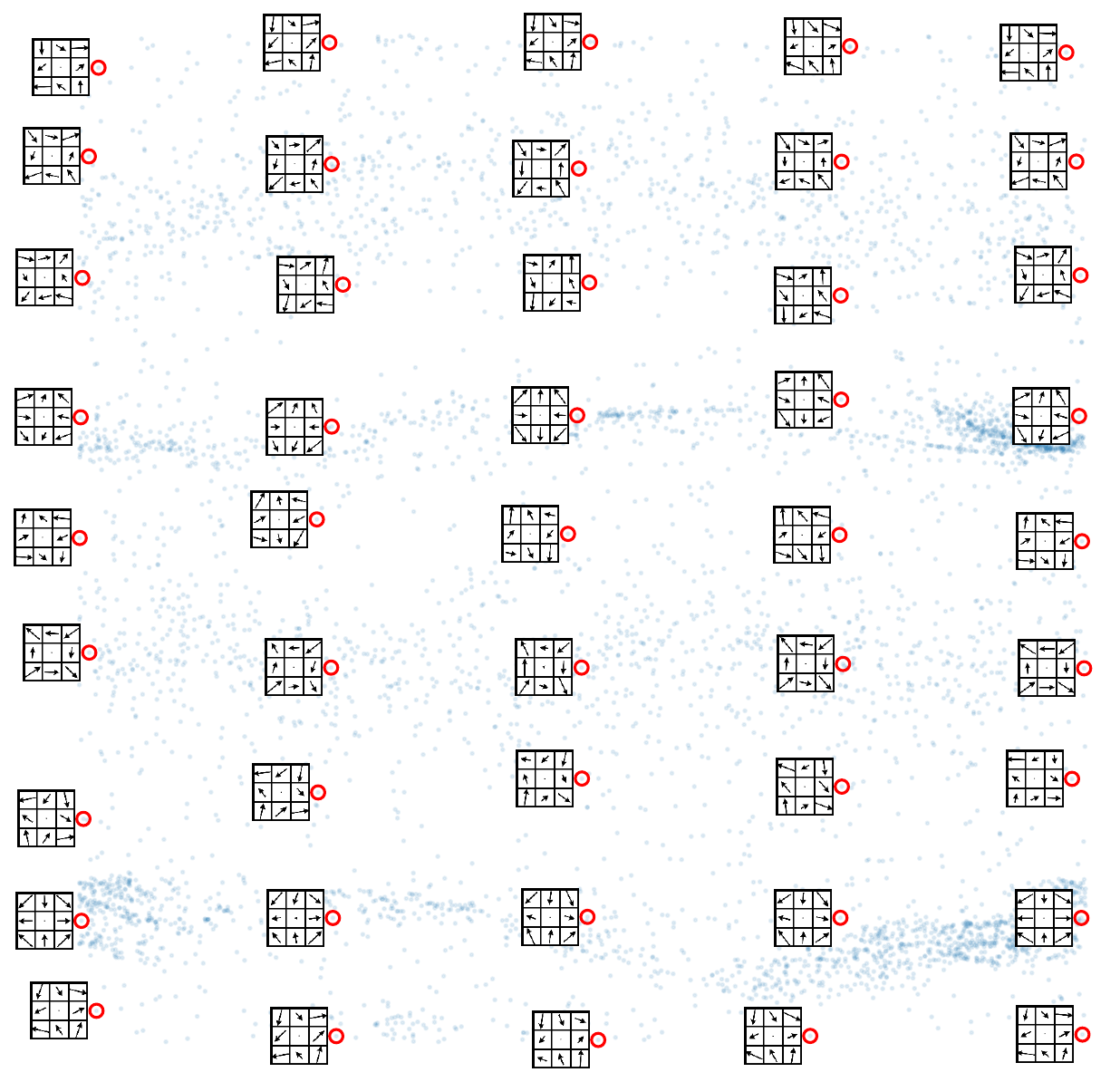}
  \caption{A sample of coordinatized flow patches from $X(1500,50)$ with directionality $r < 0.7$, arranged according to predominant direction (x-axis) and assigned fiber coordinate (y-axis). Observe that each column is roughly a copy of the low-directionality circle shown in Figure~\ref{Circle_Of_Low_Directionality}.}
  \label{low_directionality_recovered_patches}
\end{figure}

\noindent Figures~\ref{Recovered_Patch_Diagrams} and~\ref{low_directionality_recovered_patches} show samples of coordinatized patches with low- and high-directionality. Notice that each column of the low-directionality patch diagram is identical to the circle in Figure~\ref{Circle_Of_Low_Directionality}, as expected. The high-directionality patch diagram resembles Figure~\ref{Optical_Flow_Torus}, but the fiber coordinate $f$ is correlated with $\alpha - \theta\hspace{1mm} (\text{mod } 2\pi)$ rather than $\alpha$, so the flow circles rotate as one traverses the base space. This is reasonable to expect, since we know $\alpha$ cannot be used globally as a fiber coordinate (see Figure~\ref{Optical_Flow_Torus}). 

\vspace{2cm}
\subsection{Interpretation}\label{sec: Interpretation}

Our extension of the optical flow torus model from \cite{opt_flow_torus}  reflects the geometry and topology of the dense core subset $X(1500,50)$, and captures a wider diversity of frequently-occuring  patches. However, it is not clear whether the prevalence of patches with low directionality in the high-contrast data is an artifact of poor preprocessing choices or a meaningful phenomenon. 
Indeed, the interpretation of contrast norm for optical flow patches is not as straightforward as for optical range and image patches. 
In the next section, we indicate where these different high-contrast patches tend to appear in the Sintel video. \\

\vspace{2mm}
\section{Binary Step Edge Circles}\label{sec: Filaments}

In this section, we use a  finer density estimator to identify a rich network of additional dense core subsets in the space of high-contrast $3\times 3$ optical flow patches from \cite{Sintel}. 
The patches which are concentrated around this family of structures account for virtually all of the data in the top 1 percent by contrast norm, suggesting that they may be of central importance for computer vision tasks.
Furthermore, the presence of these extra structures suggests that spaces of larger optical flow patches may be distributed along a connected manifold structure which is not fully encoded by the $3 \times 3$ patches. \\

\noindent Using the same sample of $2.5\cdot 10^{5}$ high-contrast patches, we consider the dense core subset $X(50,60)$ (roughly analogous to $X(10,60)$ for the smaller sample size used in \cite{opt_flow_torus}). Figure~\ref{Distributions_Of_Directionality} shows that the distribution of directionality for $X(50,60)$ is strictly and markedly above the distributions for $X(1500, 30)$ and $X(1500,50)$ (and closer to the distribution for all of $X$). 
We henceforth   use the same predominant direction map as the feature map for our analysis, and use the same open cover $\mathcal{U}$ of $\mathbb{RP}^{1}$ as we used in Section~\ref{sec: Results}.  \\

\begin{figure}[h]
    \centering

    \begin{minipage}{0.9\textwidth}
        \centering
        \includegraphics[width=\textwidth]{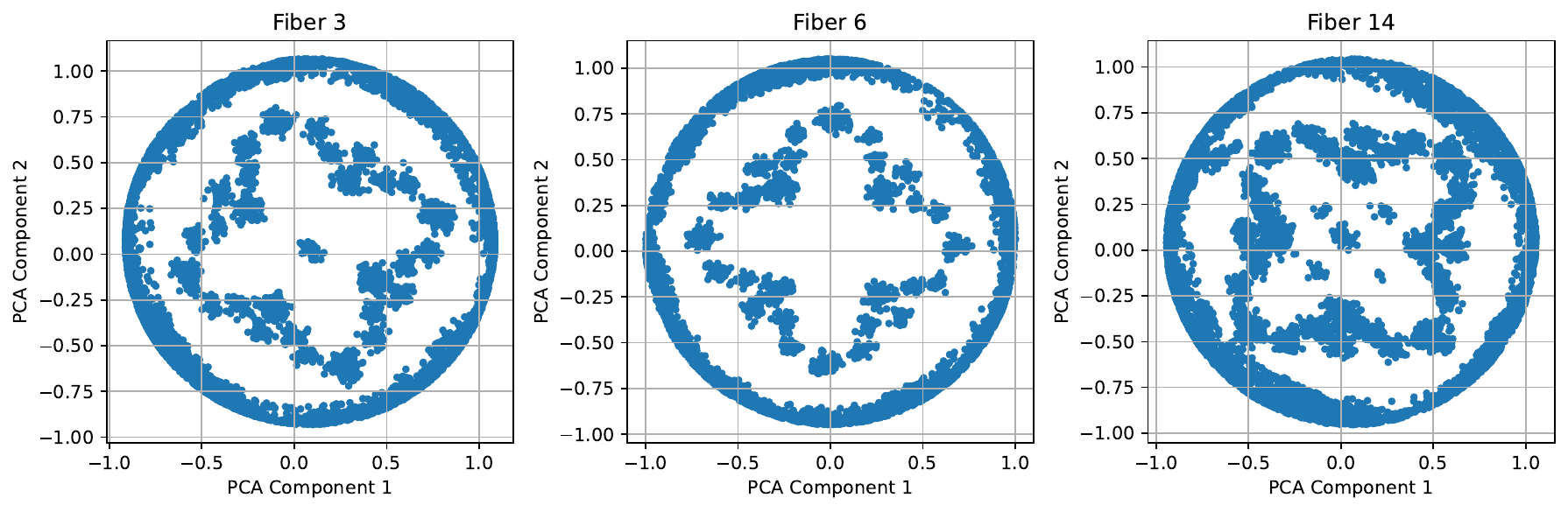}
    \end{minipage}
    \vspace{0.1cm} 
    
    \begin{minipage}{0.9\textwidth}
        \centering
        \includegraphics[width=\textwidth]{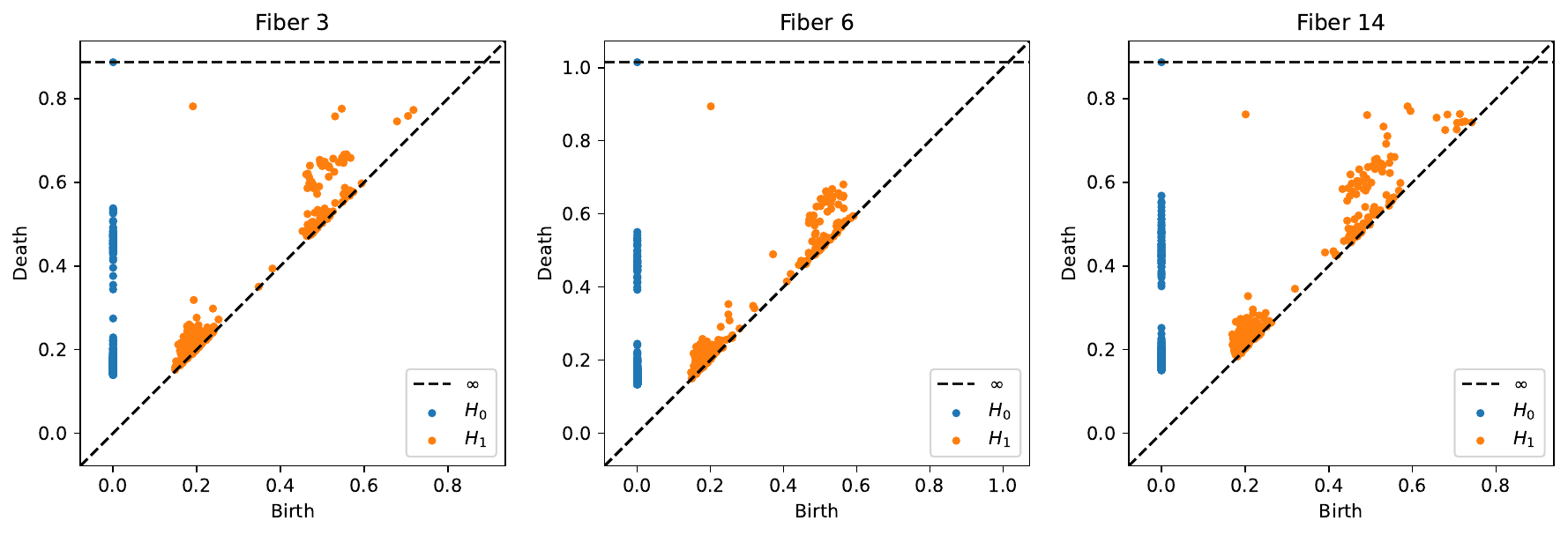}
    \end{minipage}
    \caption{ \textbf{(a)} PCA projections of several fibers of $X(50,60)$ and \textbf{(b)} Persistence diagrams of the fibers. }
    \label{Combined_Sample_Figure}
\end{figure}

\noindent Figure~\ref{Combined_Sample_Figure} shows PCA projections of several fibers $X_{j} = X\cap p^{-1}(U_{j})$ along with their persistence diagrams. In addition to a common circular feature, the projections suggest that each $X_{j}$ contains a family of disjoint filament clusters arranged in roughly the same distinguished pattern. The persistence diagrams show an interval of length scales around $\alpha = 0.3$ for which the number of connected components in each $X_{j}^{(\alpha)}$ is stable, suggesting that the 0-dimensional classes which survive beyond this range correspond to meaningful clusters. To investigate further, we conducted the following experiment:

\vspace{5mm}
\begin{enumerate}
    \item \textbf{Run Clustering On Each $X_{j}$:} Use DBSCAN~\cite{ester1996density} with $\varepsilon = 0.3$ and a minimum neighborhood size of 5 to obtain maps $\{c_{j}:X_{j}\to\mathbb{Z}\}_{j=1}^{16}$ which assign cluster labels to the points in each $X_{j}$. Let $X_{j}^{s} = c_{j}^{-1}(s)$ for each $s\in\mathbb{Z}$ and $1\leq j\leq 16$. 
    \vspace{5mm}
    
    \item \textbf{Build A Graph From The Local Clusters:} Define $X_{j}^{s}\sim X_{k}^{t}$ if and only if $X_{j}^{s}\cap X_{k}^{t}\neq \emptyset$ (in particular, $X_{j}^{s}\sim X_{k}^{t}$ implies $U_{j}\cap U_{k}\neq \emptyset)$, and let $G$ be the (undirected) graph on the vertex set $\{X_{j}^{s} \}_{j,s}$ whose edges are defined by this relation. 

    \vspace{5mm}
    \item \textbf{Compute Global Clusters:} Compute the connected components of $G$, and let $c:X\to\mathbb{Z}$ be the map which assigns to each $x\in X$  the connected component of $G$ which includes the sets containing $x$.  We interpret the level sets of $c$ as global clusters of the dataset $X$.
    
\end{enumerate}
\vspace{1cm}

\subsection{Results And Interpretation}\label{sec: Filaments Results}
\vspace{2mm} 

\begin{figure}[h]
  \centering
  \includegraphics[width=1\linewidth]{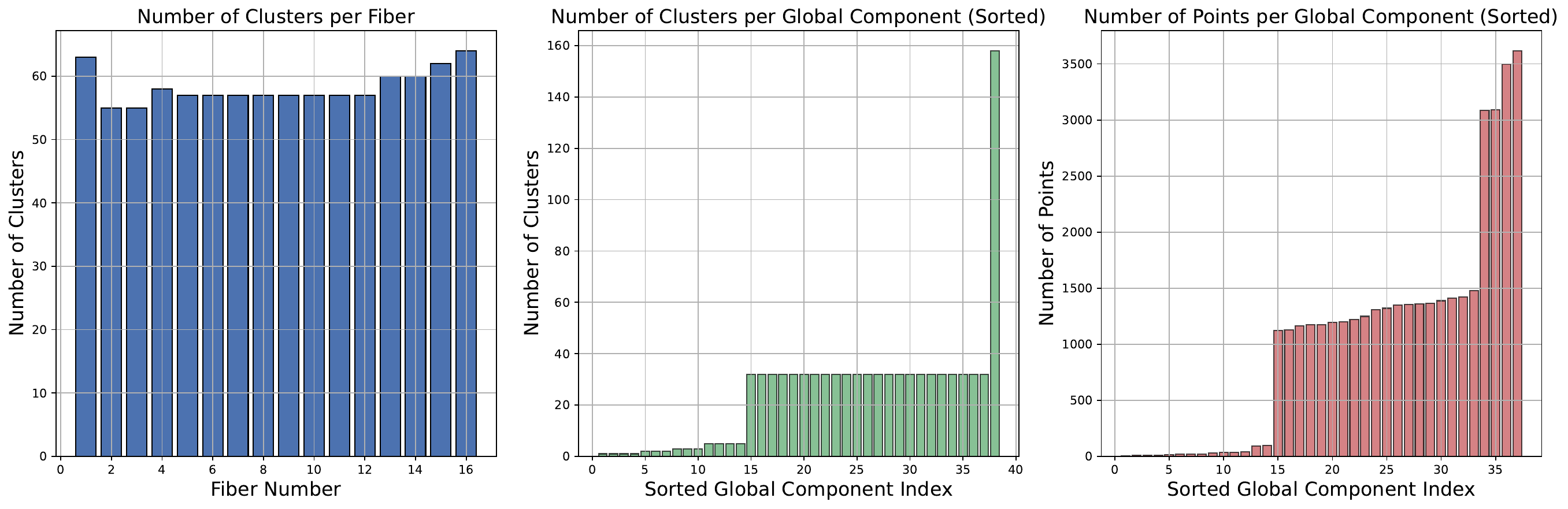}
  \caption{Results of the clustering experiment described above. Note: the cardinality of the largest component $G_{0}$ of $G$ was omitted from the rightmost plot for readability.}
  \label{Summary_Part1}
\end{figure}

\noindent Figure~\ref{Summary_Part1} shows a summary of the results of our analysis. Notice that each fiber was assigned roughly the same number of local clusters, with a mode of 57. After alignment, these clusters organized into a total of 38 global components, one of which contained the vast majority of the data. We denote the associated component of $G$ by $G_{0}$. Of the remaining components, there were 23 secondary components comprised of exactly 32 local clusters each -- twice the number of open sets in our cover. These components contained over 90 percent of the data unaccounted for by the largest component. Among them, the four largest (by cardinality) were approximately the same size, and the others were each roughly one third as a large. \\ 

\begin{figure}[h]
  \centering
  \includegraphics[width=\linewidth]{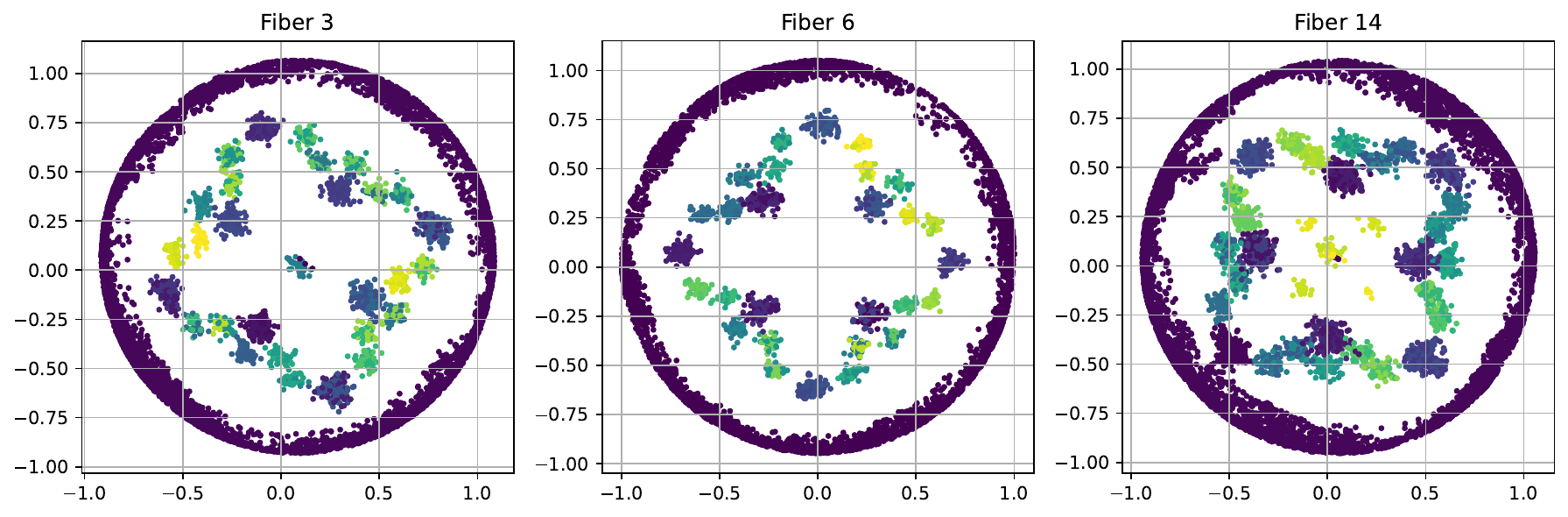}
  \caption{ }
  \label{Sample_Clustered_Fibers}
\end{figure}

\vspace{5mm}
\noindent\textbf{Structure Of The Fibers:} Figure~\ref{Sample_Clustered_Fibers} shows the 2D projections of several $X_{j}$ colored according to the cluster labels assigned by DBSCAN. Notice that each apparent 'filament' in the 2D projections is actually composed of two distinct clusters centered around antipodal points on $\mathbb{S}^{15}$ (see Figure~\ref{Optical_Flow_Annulus}).\\

\begin{figure}[h!]
    \centering

    \begin{minipage}{0.7\textwidth}
        \centering
        \includegraphics[width=\textwidth]{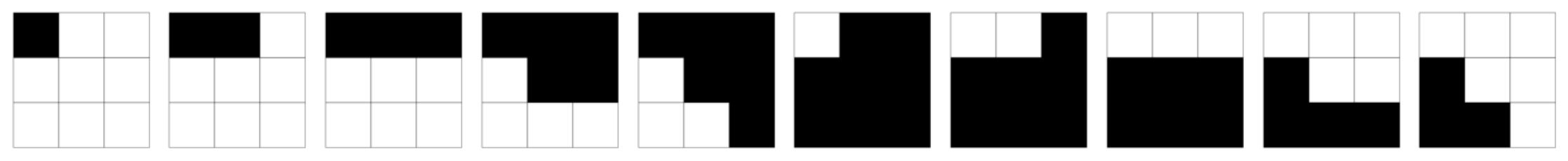}
    \end{minipage}
    
    \vspace{0.1cm} 
    
    \begin{minipage}{0.7\textwidth}
        \centering
        \includegraphics[width=\textwidth]{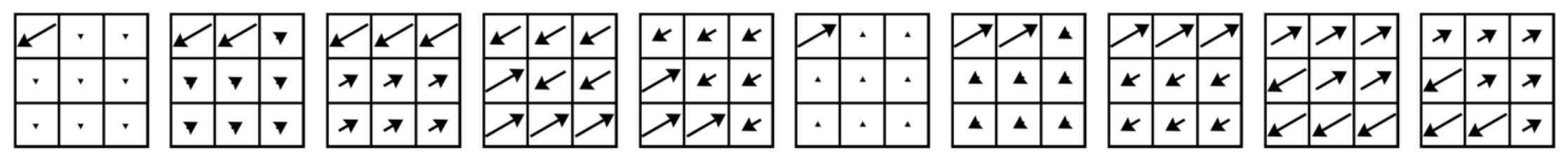}
    \end{minipage}

\vspace{0.1cm} 
    
    \begin{minipage}{0.7\textwidth}
        \centering
        \includegraphics[width=\textwidth]{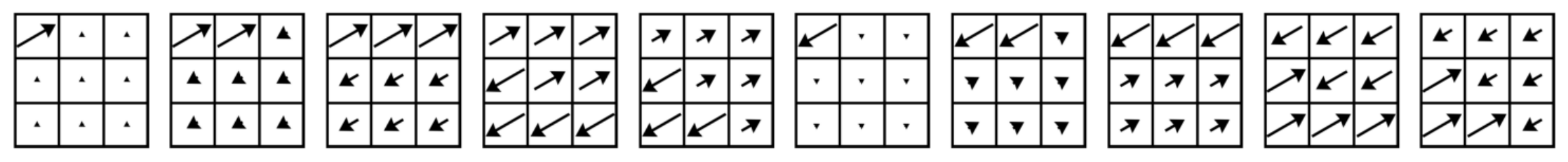}
    \end{minipage}

    \caption{Two ways to identify binary step-edge flow patches in the fiber at $\theta = \frac{\pi}{6}$ with the corresponding binary step-edge range patches}
    \label{Binary_Patch_Sample}
\end{figure}

\noindent Figure~\ref{Binary_Patch_Sample} shows representative patches from some of the local filament clusters in one of the fibers. Observe that each cluster corresponds to one of the 56 possible $3 \times 3$ binary step-edge range patches with camera panning applied along the predominant flow axis (see Appendix~\ref{sec: Step-Edge Patches} for a list of all the binary step-edge range patches). On the other hand, there are two distinct ways to make the identification; if we choose an orientation $\hat{n}\in\mathbb{S}^{1}$ for the predominant direction (axis) in $\mathbb{RP}^{1}$ associated with the fiber, we obtain the mean patch in each filament cluster by applying a camera translation in the direction $\hat{n}$ to one of the binary step-edge range patches (and then normalizing). If we choose the opposite orientation, each filament cluster is identified with the opposite binary range patch. In any case, notice these patches have high directionality, so predominant direction is a meaningful measurement. \\

\begin{figure}[h!]
  \centering
  \includegraphics[width=\linewidth]{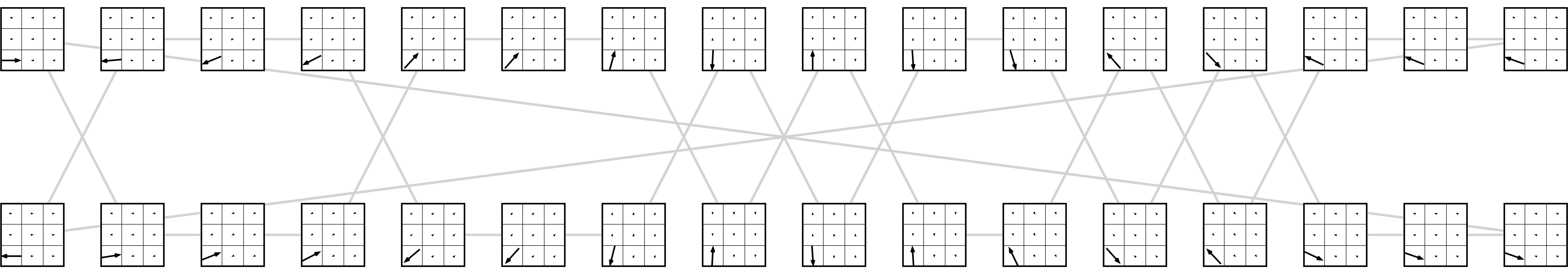}
  \caption{A visualization of a connected component of $G$. Each node is labeled by a patch from the corresponding local cluster; edges representing intersections between local clusters are shown. Patches in the same column are represent local clusters over the same open set in the cover $\{U_{j}\}_{j=1}^{16}$ of $\mathbb{RP}^{1}$.}
  \label{Double_Cover_Patches}
\end{figure}

\noindent\textbf{Global Structure:} Figure~\ref{Double_Cover_Patches} shows representative patches from each of the 32 local clusters which comprise $G_{1}$, one of the larger secondary components of $G$. Notice how each patch has the same single 'active' component flow vector, and that the direction of this vector rotates as one travels along $G_{1}$. As the active vector completes a full rotation, $G_{1}$ traverses the base space twice, intersecting each fiber $X_j$ in two distinct places and forming a double cover of $\mathbb{RP}^{1}$.

\begin{figure}[h!]
  \centering
  \includegraphics[width=0.9\linewidth]{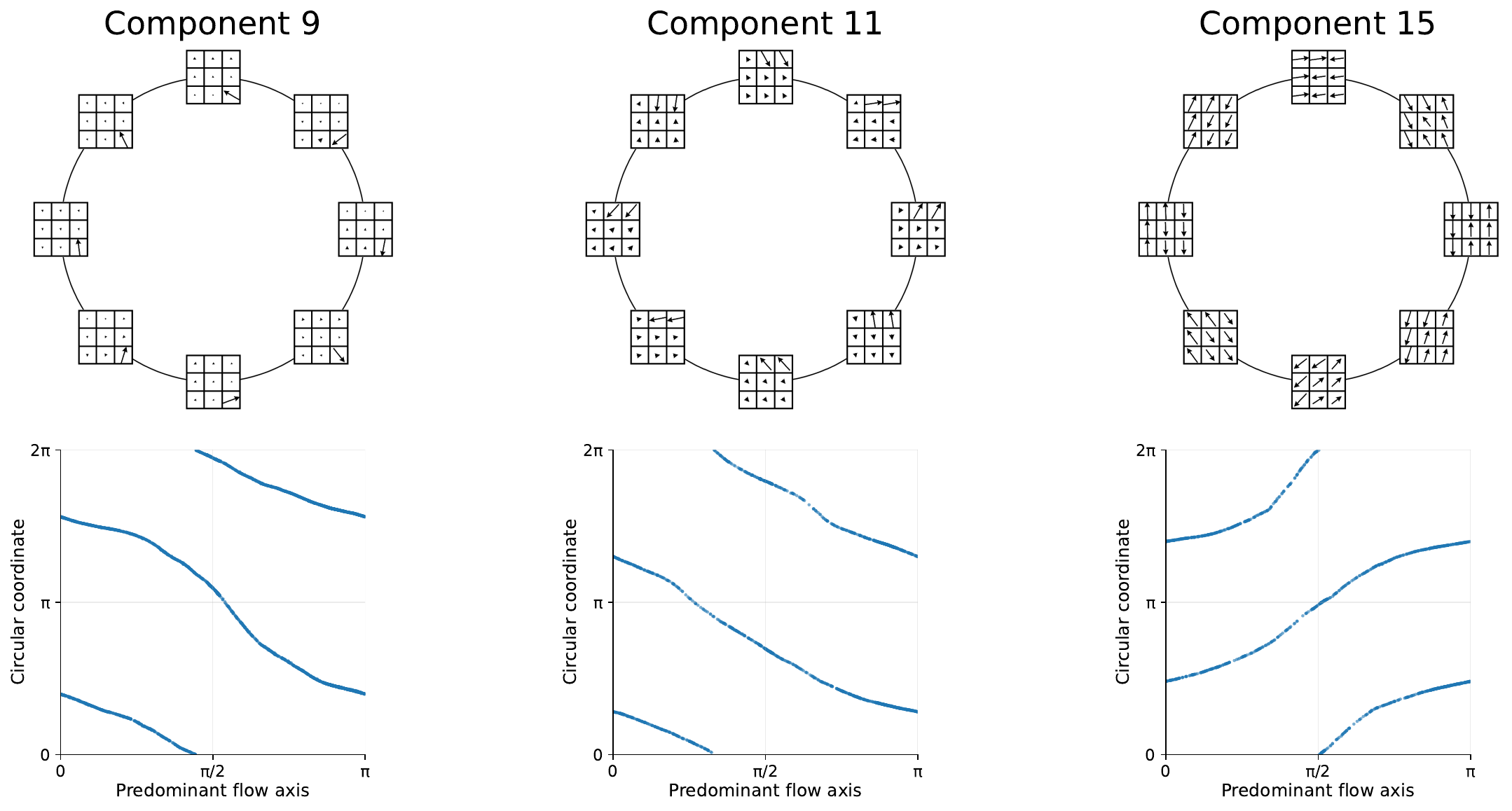}
  \caption{Exemplary patches from three global connected components of $G$, placed in $\mathbb{S}^{1}$ according the computed circular coordinate.}
  \label{Sample_CCs}
\end{figure}

\noindent In light of these results, one would expect that each of the 23 secondary components of $G$ had a single prominent circular feature, and indeed this was the case. We verified this with persistent homology and used the sparse circular coordinates algorithm to parameterize each one. The results for several components are shown in Figure~\ref{Sample_CCs}. 

\vspace{5mm}

\noindent\textbf{Missing Circles: }Each of the 23 secondary components of $G$ corresponds to a pair of binary range patches with camera motions applied along all possible directions. This suggests that there should be a total of 28. On the other hand, Figure~\ref{Summary_Part1} shows that $G_{0}$ is comprised of 158 local clusters. Since the extended optical flow torus is connected, we expected the largest connected component of $G$ (by cardinality) to contain roughly a single local cluster in each fiber. \\

\begin{figure}[h!]
    \centering
    \begin{minipage}[t]{0.38\textwidth}
        \centering
        \vspace{0pt} 
        \includegraphics[width=\linewidth]{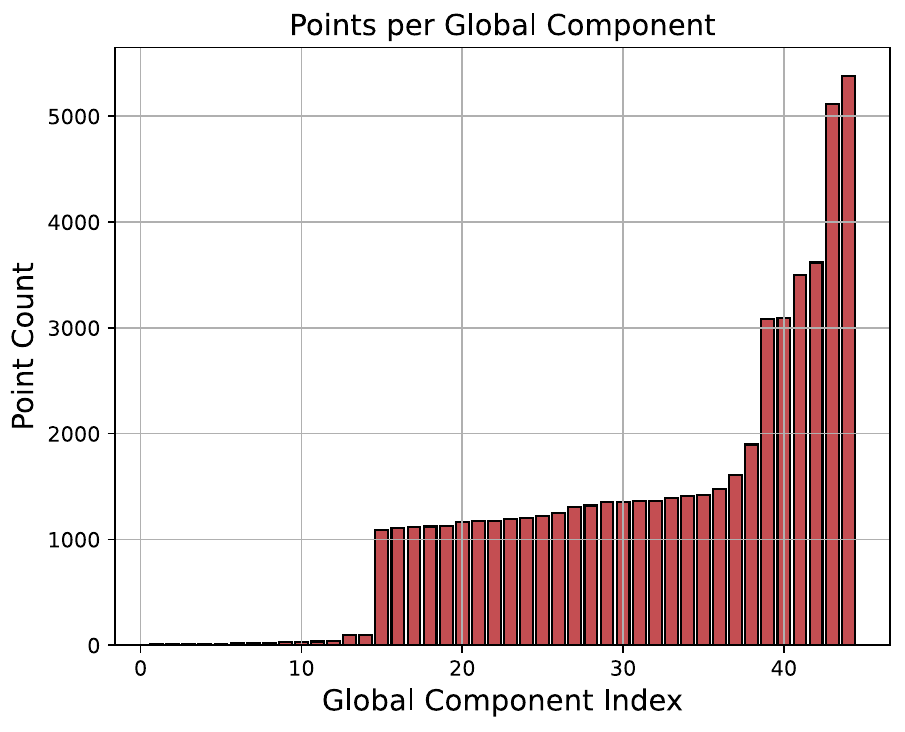}
        \vspace{0.6em}

        \includegraphics[width=\linewidth]{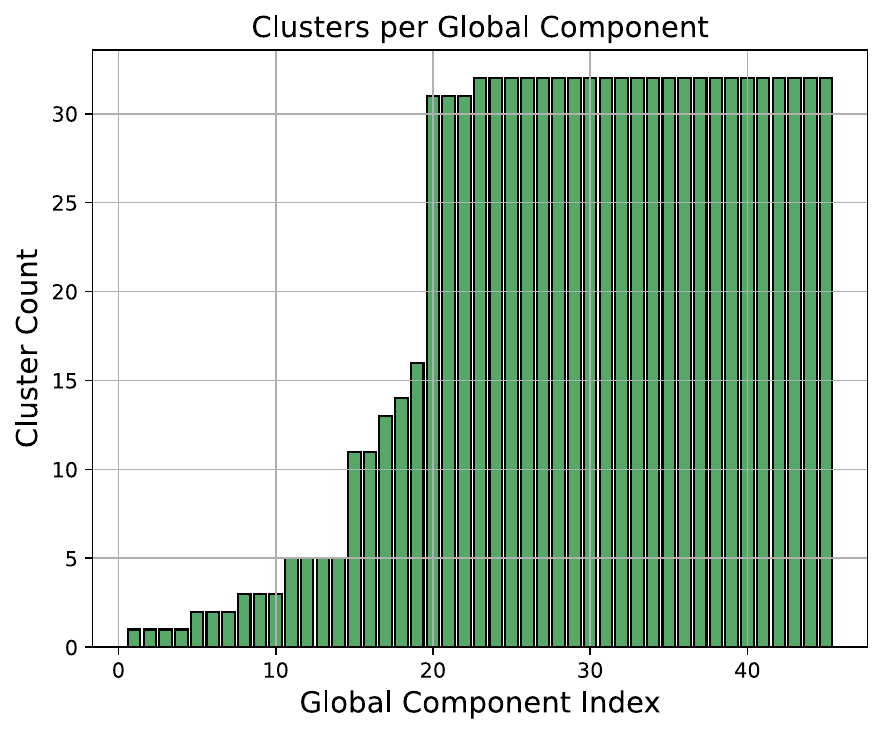}
        \vspace{0.2em}
        {\small\textbf{(a)}}
    \end{minipage}
    \hspace{4mm}
    \begin{minipage}[t]{0.55\textwidth}
        \centering
        \vspace{10pt} 
        \includegraphics[width=\linewidth]{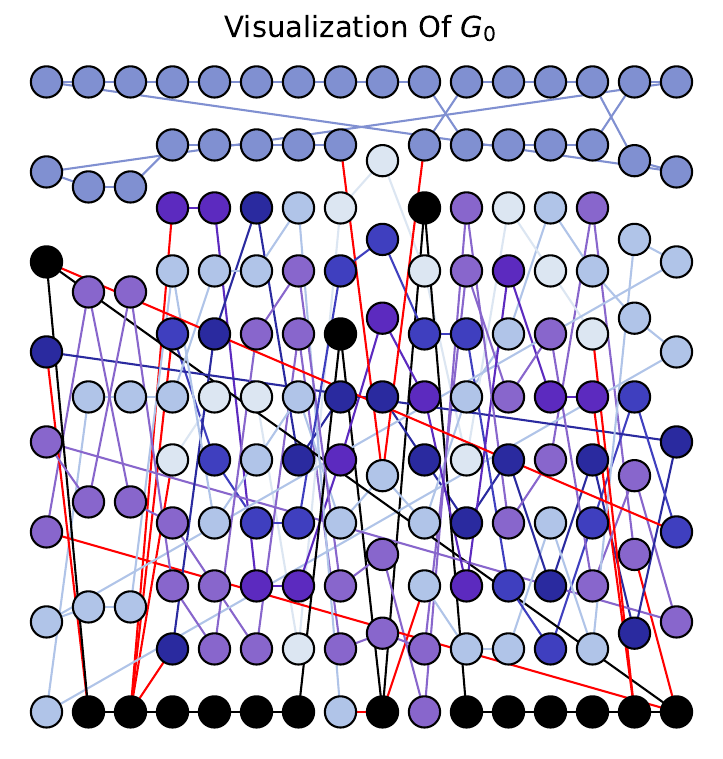}
        \vspace{0.2em}
        {\small\textbf{(b)}}
    \end{minipage}
    \caption{\textbf{(a)} A summary of the connected components of $G^{(0.07)}$. \textbf{(b)} A visualization of the largest connected component of $G$. Each node represents a local cluster, and each edge represents an intersection. The nodes are colored according to their connected components in the complete subgraph $G^{(0.07)}$. Nodes corresponding to data from the (extended) optical flow torus are shown in black. Edges which do not appear in $G^{(0.07)}$ are shown in red.}
    \label{Summary_Part2}
\end{figure}

\noindent With further investigation, we found that in each fiber, $G_{0}$ contained a single large local cluster and approximately 10 much smaller local clusters. In fact, the local clusters appeared to organize into nearly-distinct sub-components, connected only in a few places (see Figure~\ref{Summary_Part2}b). This clearly suggested that our 'missing' circles were tangled up with the extended flow torus. \\
 
\vspace{3mm}

\noindent \textbf{Removing Edges: }To untangle the various pieces of the main component, we assigned a weight to each edge of $G$ according to the formula

\begin{equation}
    w(e_{jk}) = \frac{|X_{j}\cap X_{k}|}{\text{max}\left(|X_{j}|,|X_{k}|\right)}
\end{equation}
\vspace{1mm}

\noindent Notice that edges which correspond to intersections with only a few points are assigned small weights. Edges corresponding to intersections of local clusters with vastly different sizes are also assigned small weights, since the numerator in the formula above is bounded above by the cardinality of the smaller set. \\

\begin{figure}[h!]
    \centering
    \includegraphics[width=0.5\textwidth]{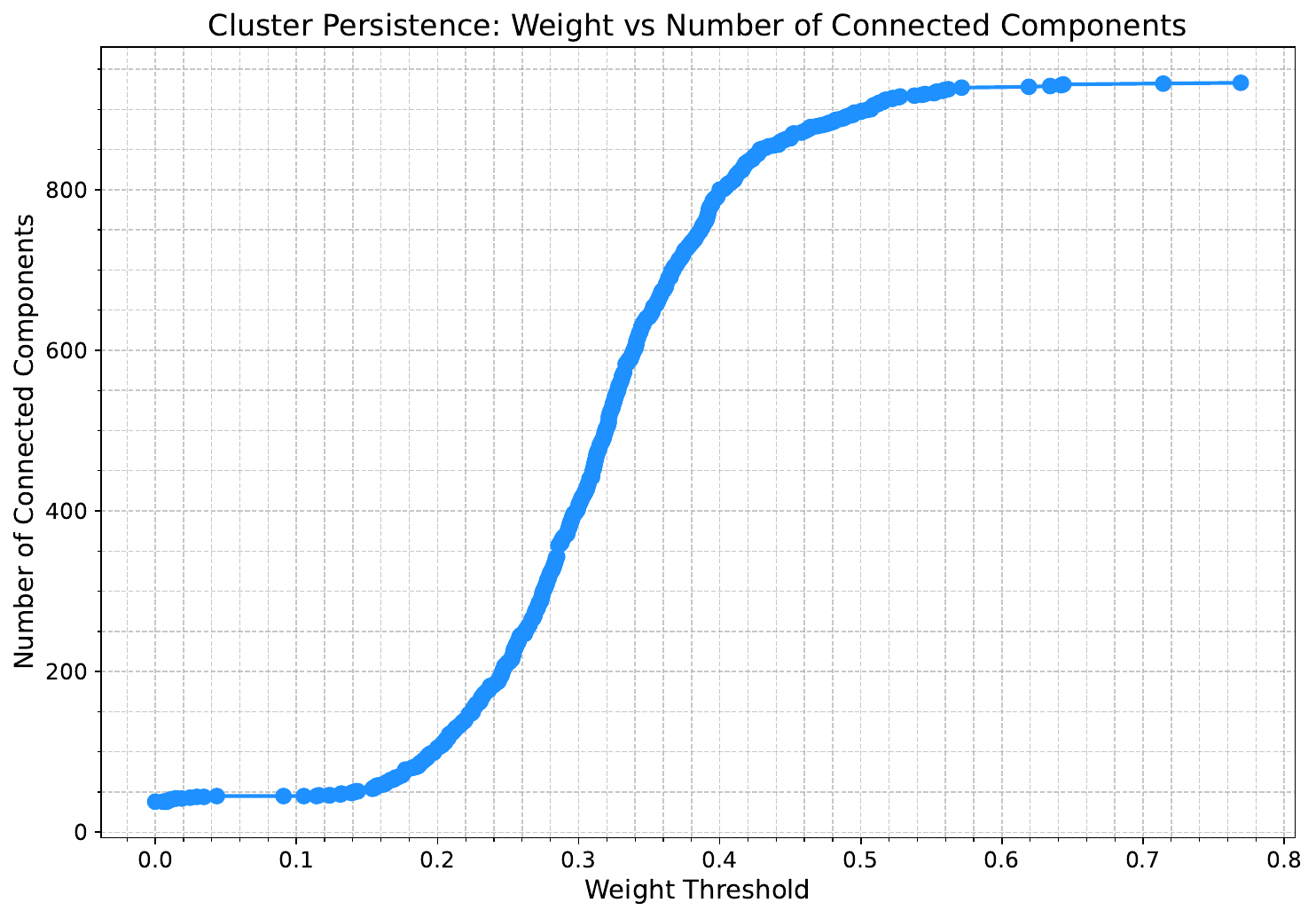}
    \caption{ }
    \label{Cluster_Persistence}
\end{figure}

\noindent Defining $w(v_{j}) = \infty$ for each node $v_{j}$, we obtained a filtration of $G$ by $\{G^{(t)} = w^{-1}(t,\infty]\}$. Figure~\ref{Cluster_Persistence} shows how the number of connected components of $G^{(t)}$ varied with $t$. Observe that for small $t$, the number of connected components was fairly stable. In particular, there was an interval of weights around $0.07$ over which the number of components did not change, possibly indicating that edges with weights below this threshold were caused by noise. \\

\noindent After removing the 14 edges of $G$ with weight below $0.07$, the remaining graph $G^{(0.07)}$ had 45 components, 27 of which had a 1-dimensional persistence class suitable for use in the sparse circular coordinates algorithm. We verified that these components corresponded to the extended optical flow torus and 26 of the 28 anticipated binary step-edge circles. Unlike for $G$, the primary component of $G^{(0.07)}$ by cardinality did not comprise an exceptionally large number of local clusters (see~\ref{Summary_Part2}a) -- in fact, it contained exactly one in each fiber, represented by the black nodes in Figure~\ref{Summary_Part2}b).\\

\noindent Note that we recovered three of the five 'missing circles', two of which were both notably larger in cardinality than the other circular components (though nowhere near the size of the primary component). These larger circles corresponded to the horizontal linear binary step-edge patches.  \\
\vspace{5mm}

\begin{figure}[h!]
    \centering
    
    \begin{minipage}[t]{0.48\textwidth}
        \centering
        \vspace{0pt}
        \includegraphics[width=\linewidth]{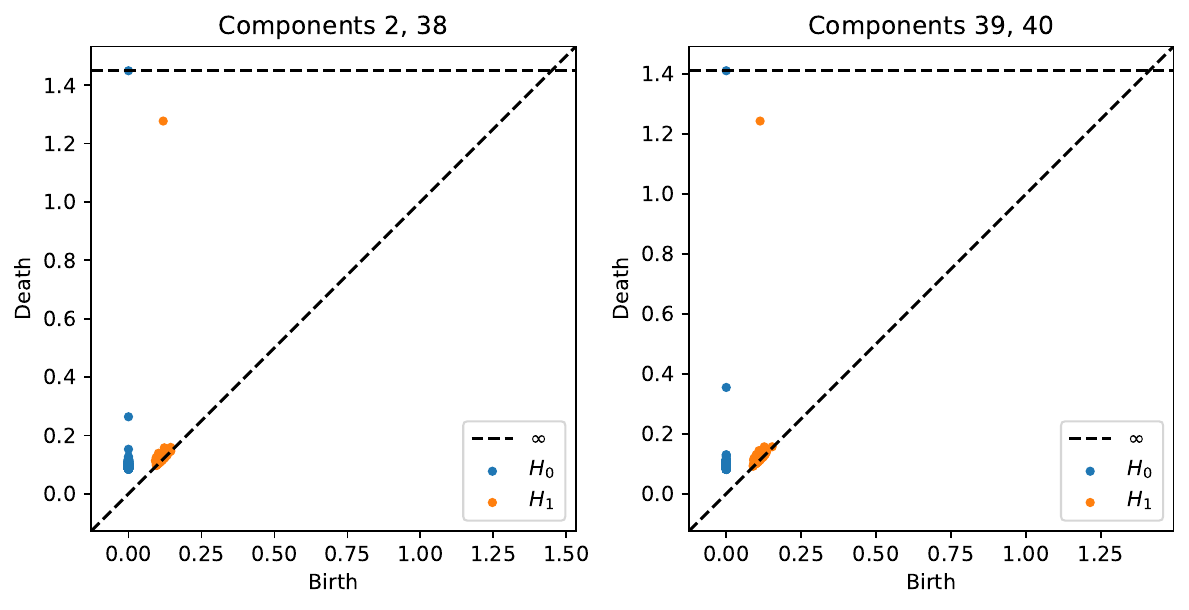}
    \end{minipage}
    \hfill
    \begin{minipage}[t]{0.48\textwidth}
        \centering
        \vspace{0pt}
        \includegraphics[width=\linewidth]{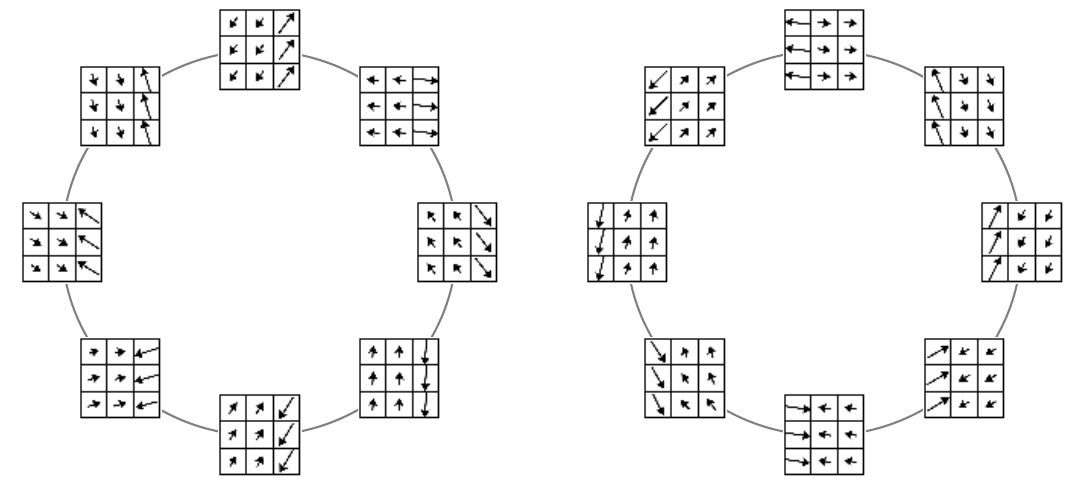}
    \end{minipage}

    \caption{
    (a) The persistence diagrams for the unions of component pairs $\{2, 38\}$ and $\{39, 40\}$ of $G^{(0.07)}$, respectively.
    Each union is disconnected until the components combine to form a single loop at a length scale of approximately $0.3$.
    (b) Sample patches from each component pair arranged by assigned circular coordinate.
    }
    \label{Missing_Circles}
\end{figure}

\noindent\textbf{Circle Fragments:} None of the remaining 18 connected components of $G^{(0.07)}$ exhibited a 1-dimensional persistence class usable as input for the sparse circular coordinates algorithm. On the other hand, we found four components which were comparable in cardinality to some of the smaller circular components, and which contained more than 90 percent of the data still unaccounted for. Figure~\ref{Missing_Circles}a shows the persistence diagrams for the unions of the paired-off components, clearly suggesting that each of these four components made up roughly half of one of the missing circles. To confirm this, we used the sparse circular coordinates algorithm to parameterize each pair -- the results are shown in Figure~\ref{Missing_Circles}b.  \\          

\vspace{5mm}
\noindent\textbf{Outlier Patches: }Together, the primary component (i.e., the extended flow torus) and filament circles account for virtually all of $X(50,60)$. From the projections in Figure~\ref{Sample_Clustered_Fibers}, we see that the few outlier patches in the data are also organized in a distinguished pattern of local clusters, though these extra clusters are only present in some of the more densely populated fibers.\\

\begin{figure}[h!]
  \centering
  \includegraphics[width=0.5\linewidth]{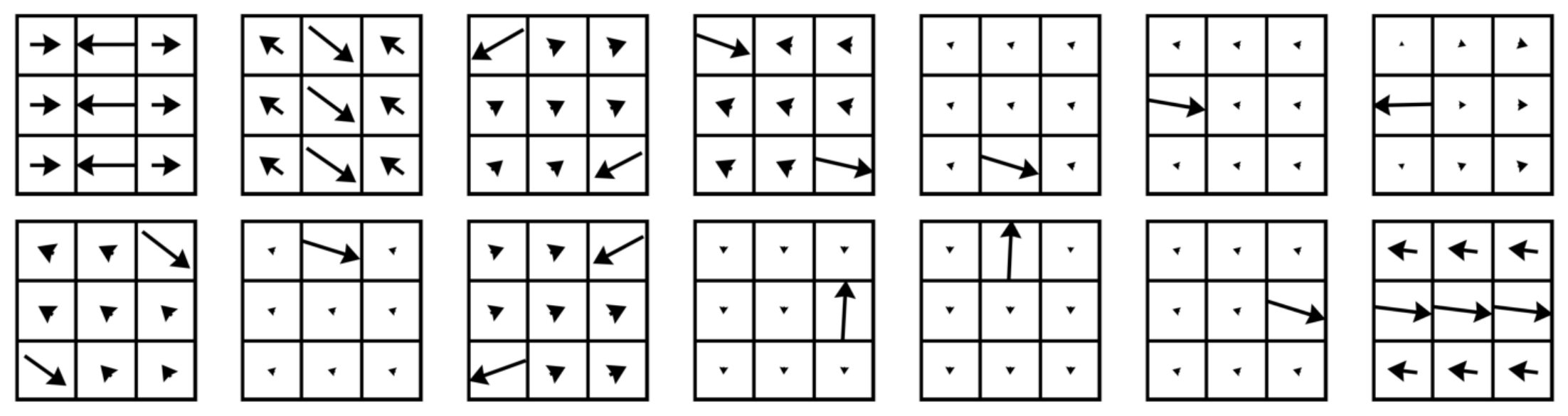}
  \caption{ }
  \label{fig:Quadratic_Patches}
\end{figure}

\noindent Figure~\ref{fig:Quadratic_Patches} shows a representative sample of patches from the outlier clusters. The patches with only one or two active components apparently correspond to rare/anomalous binary range patches (with camera motion applied along one of the more common predominant directions). In principle, one could conceive of full circular components corresponding to pairs of these patches, just as we saw for the other filaments. The remaining outliers correspond to range patches with quadratic gradients. In particular, these patches take the form 

\vspace{0.5mm}
\begin{equation*}
    P(s,t) = \text{cos}(t)\left(\text{cos}(s)e_{3}^{u} + \text{sin}(s)e_{4}^{u}\right) + \text{sin}(t)\left(\text{cos}(s)e_{3}^{v} + \text{sin}(s)e_{4}^{v}\right)
\end{equation*}
\vspace{0.5mm}

\noindent where $e_{3}^{u},e_{4}^{u},e_{3}^{v},e_{3}^{v}$ denote horizontal and vertical optical flow DCT basis vectors (see Figure~\ref{Flow_DCT}). High-contrast optical image patches with quadratic gradients are central to the Klein bottle model described in \cite{Klein_bottle}. Here again, one could conceive of a full secondary optical flow torus parameterized as above, or even a family of optical flow tori interpolating between the linear and quadratic gradients. \\
 
\vspace{2mm}
\subsection{Interpretation}\label{sec: Filament Interpretation}
\begin{figure}[h]
  \centering
  \includegraphics[width=\linewidth]{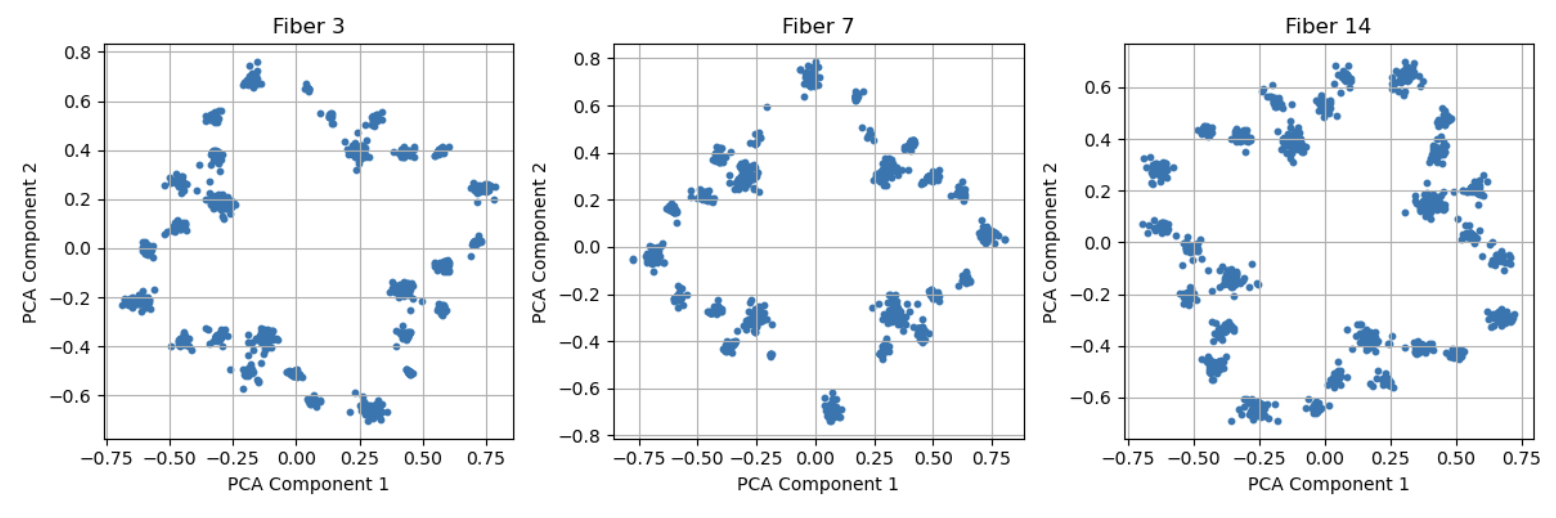}
  \caption{PCA projections of fibers of $\tilde{X}(30,80)$ (where $\tilde{X}$ denotes the top 5 percent of patches in our high-contrast sample $X$. Hence $\tilde{X}$ represents the top 1 percent of all optical flow patches by contrast norm).}
  \label{fig:HC1_Sample_Projections}
\end{figure}

\vspace{2mm}
Our findings show that a dense core subset of the space of 3 x 3 high-contrast optical flow patches is concentrated around a family of circles, each corresponding to a pair of binary step-edge range patches with camera motions applied. The original contrast norms of the patches in these filament circles are generally much higher than the contrast norms of patches near the extended optical flow torus. In fact, if we restrict our attention to the top 1 percent of patches by contrast norm, nearly \textit{all} of the data is concentrated around these filaments circles -- see Figure~\ref{fig:HC1_Sample_Projections}. This is intuitively reasonable, as the binary patches will inevitably tend to have higher contrast norm than the linear gradient patches of the main flow torus. \\

\begin{figure}[h!]
    \centering
    \begin{minipage}[t]{0.48\textwidth}
        \centering
        \vspace{0pt}
        \includegraphics[width=\linewidth]{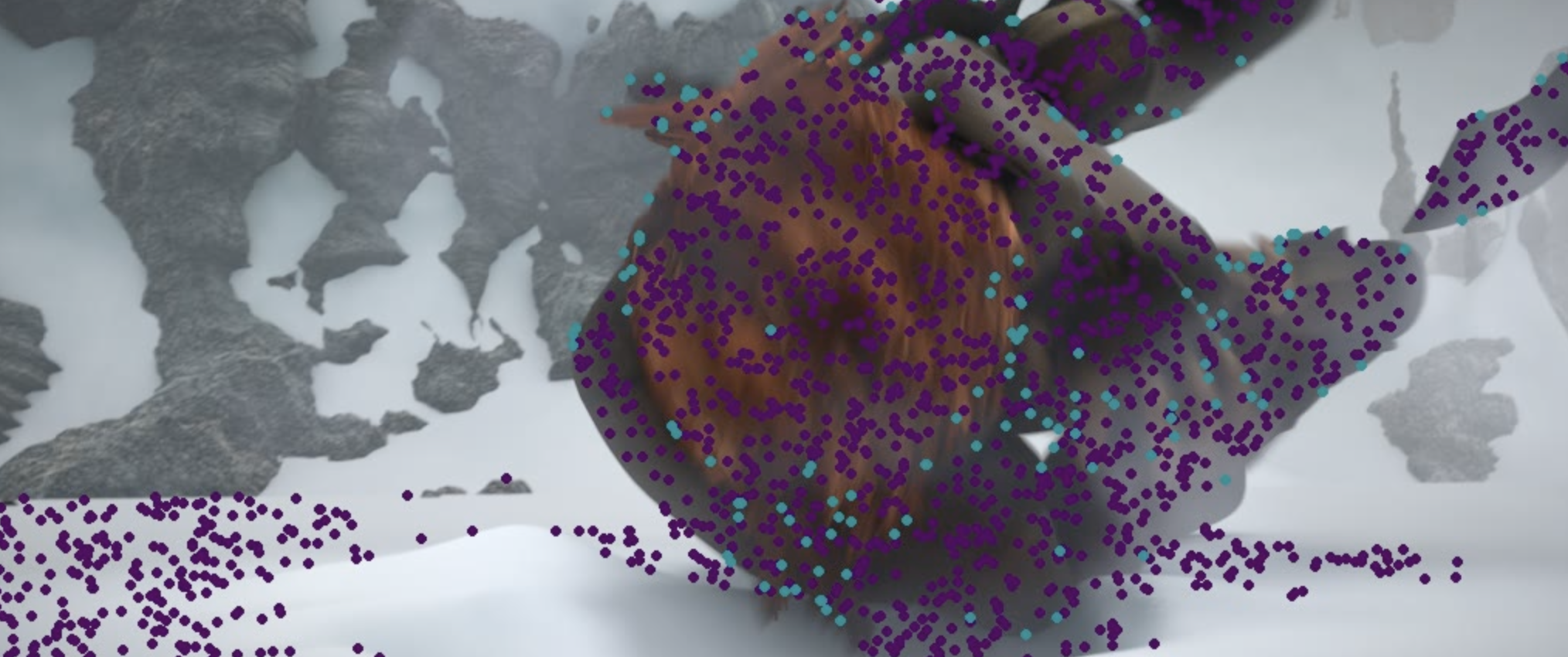}
    \end{minipage}
    \hfill
    \begin{minipage}[t]{0.48\textwidth}
        \centering
        \vspace{0pt}
        \includegraphics[width=\linewidth]{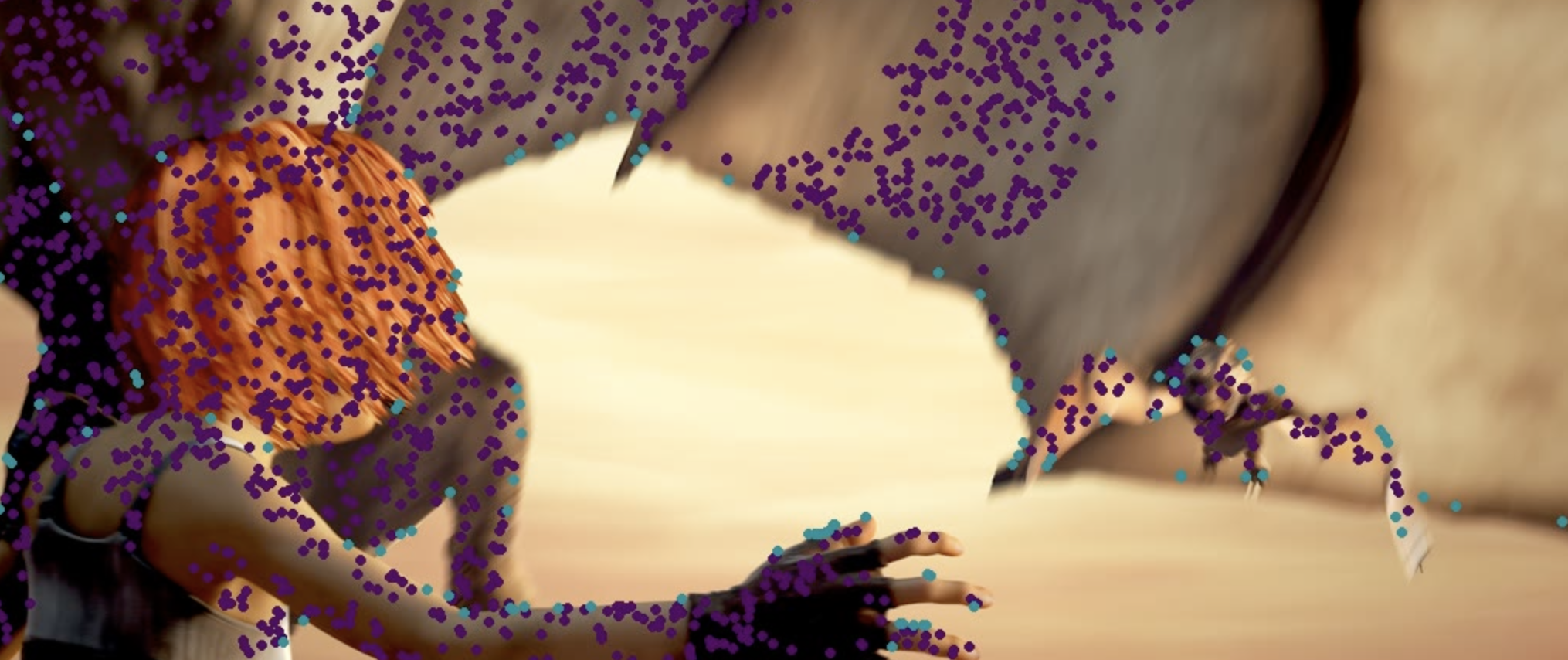}
    \end{minipage}

    \vspace{0.8em}
    \begin{minipage}[t]{0.48\textwidth}
        \centering
        \vspace{0pt}
        \includegraphics[width=\linewidth]{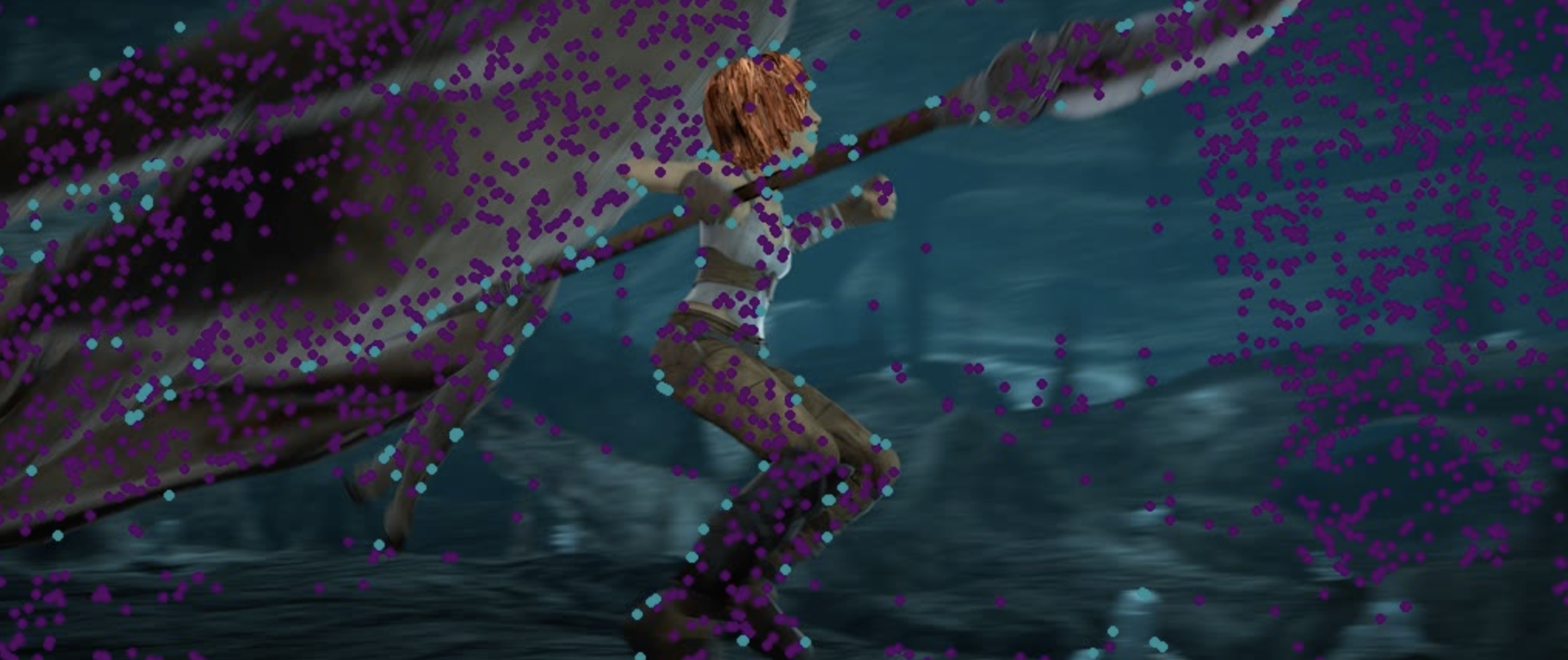}
    \end{minipage}

    \caption{Three frames from the Sintel video with dots labeling the original locations of high contrast patches from our sample. Dark purple dots indicate patches in the top 20 percent by contrast norm; light blue dots indicate patches in the top 1 percent.}
    \label{Labeled_Frames}
\end{figure}

\noindent Anecdotally, we found that optical flow patches in the top 1 percent by contrast norm tend to appear at motion boundaries, whereas patches in the top 20 percent can be found near moving bodies of high image contrast (such as hair or textured surfaces -- see Figure~\ref{Labeled_Frames}). This suggests that the binary flow patches may be of particular relevance for computer vision tasks. \\  

\begin{figure}[h!]
  \centering
  \includegraphics[width=0.75\linewidth]{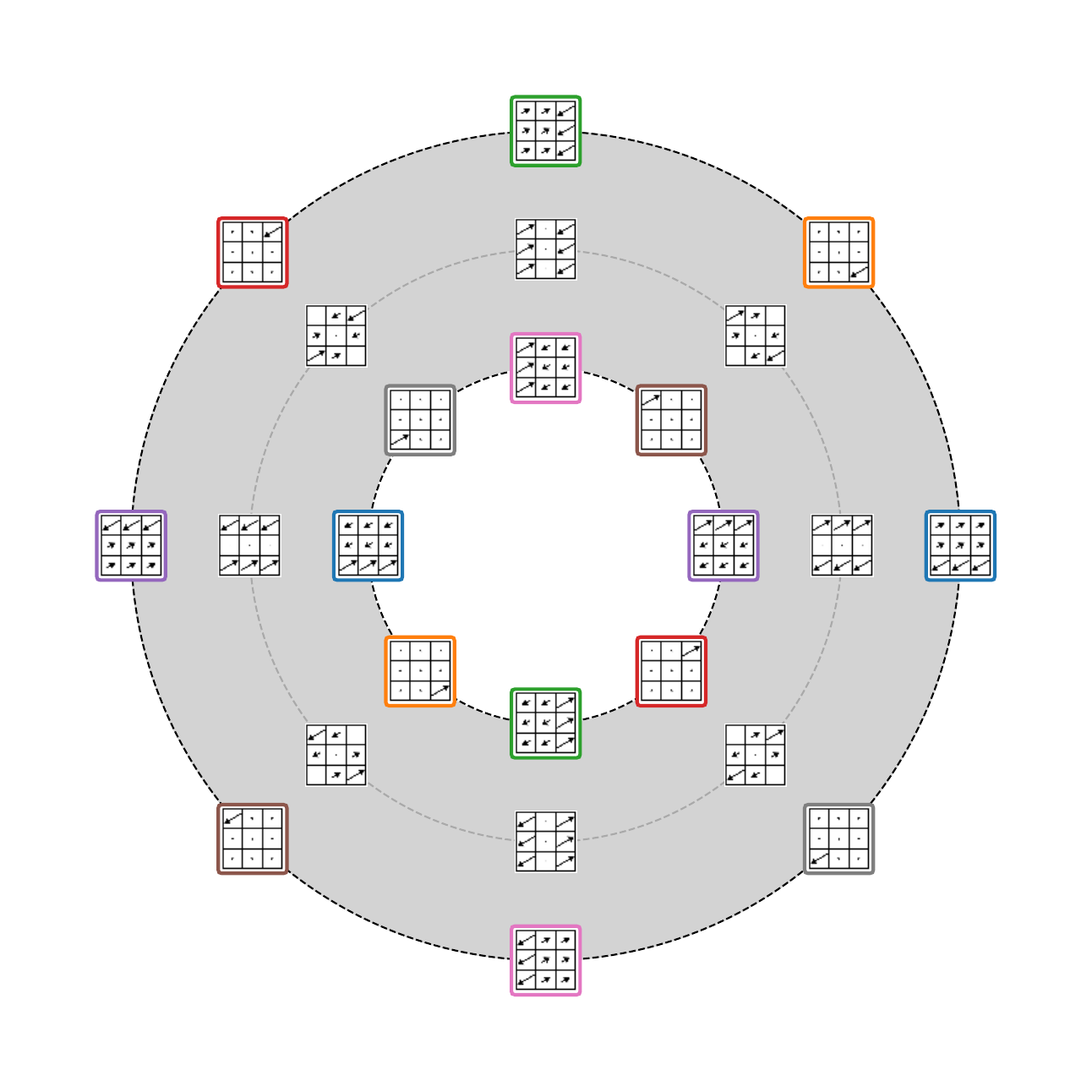}
  \caption{}
  \label{Optical_Flow_Annulus}
\end{figure}

\noindent\textbf{Connected Model:} Figure~\ref{Optical_Flow_Annulus} shows the layout of a single fiber in our expanded model of the 3 x 3 high-contrast optical flow dataset (excluding patches near the main flow torus with low directionality). The central circle represents a slice of the flow torus, and some of the binary flow patches are shown on the boundary circles (which are antipodal in $\mathbb{S}^{15}$ -- hence they appeared superimposed in the PCA projections). Binary patches belonging to the same global circular cluster in the dataset are indicated by borders with matching colors. 

\noindent Our analysis showed that only the central flow circle and small regions around each binary patch are dense.  However, by analogy with \cite{Lee_Mumford}, one could  conceive of a single continuous manifold structure which likely emerges for patches of larger size -- a parameterized family of annuli obtained by applying uniform camera panning to patches from the step-edge annulus model for range patches (see Section~\ref{sec: Prior Work}). This manifold $\mathcal{M}$ is homotopy equivalent to the optical flow torus, though $(\mathcal{M},p, \mathbb{RP}^{1})$ is topologically nontrivial as a fiber bundle with annular fiber. Equivalently, one could view $\mathcal{M}$ as the total space of the disk bundle of a non-orientable line bundle over the Adams flow torus $\mathcal{T}$. Just as the boundary of the Mobius band is a single circle, the boundary of $\mathcal{M}$ consists of a single torus, and each fiber of $p|_{\partial \mathcal{M}}$ is a \textit{pair} of circles; as indicated in Figure~\ref{Optical_Flow_Annulus}, the binary step edge patches are concentrated near $\partial\mathcal{M}$, and as one traverses the base space, the boundary circles in each fiber are interchanged. \\

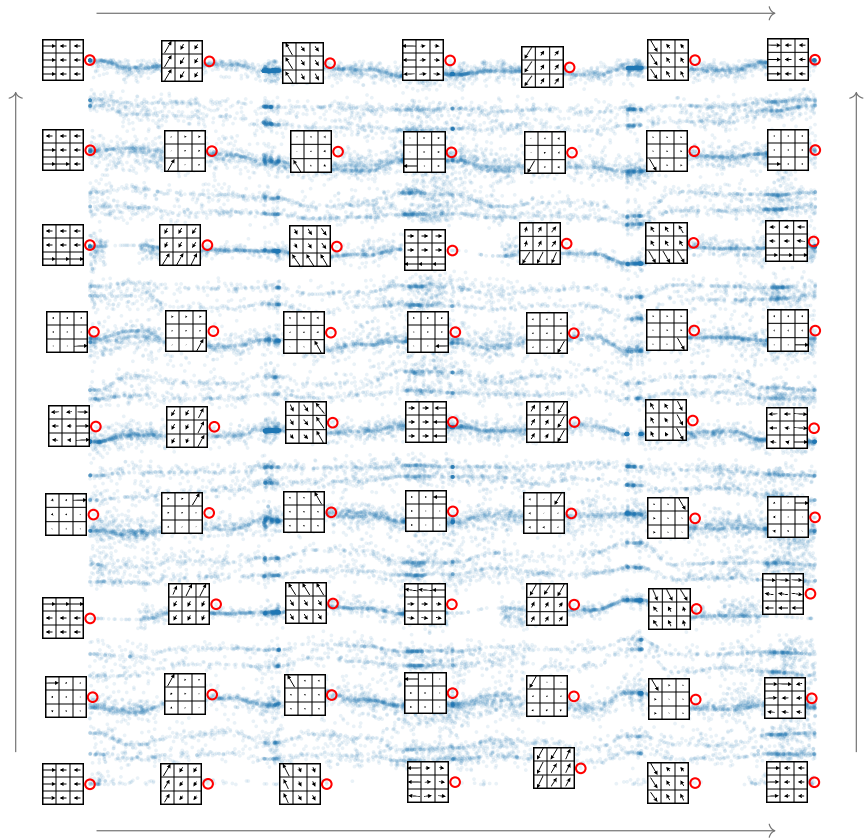
\begin{figure}[h!]
\centering
\begin{tikzpicture}
  \node[anchor=south west, inner sep=0] (img) at (0,0)
    {\includegraphics[width=0.65\linewidth]{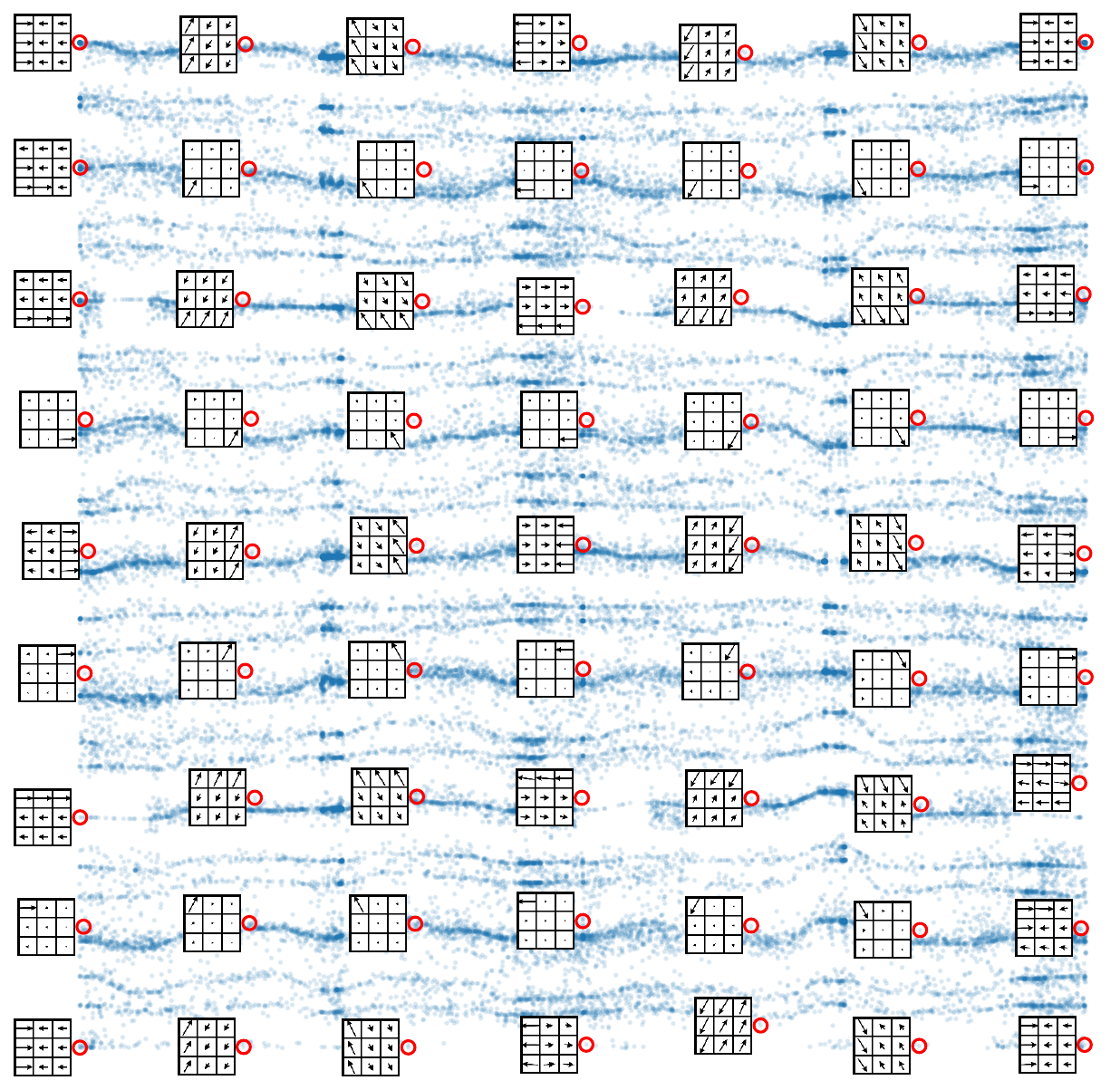}};
  \begin{scope}[x={(img.south east)}, y={(img.north west)}]

    \tikzset{torusarrow/.style={->, line width=0.45pt, draw=black!55}}

    \draw[torusarrow] (0.08,1.02) -- (0.92,1.02);
    \draw[torusarrow] (0.08,-0.02) -- (0.92,-0.02);

    \draw[torusarrow] (-0.02,0.08) -- (-0.02,0.92);
    \draw[torusarrow] (1.02,0.08) -- (1.02,0.92);

  \end{scope}
\end{tikzpicture}
\caption{A sample of binary step-edge patches from $X_{\text{step}}$ arranged according to lifted predominant direction ($x$-axis) and assigned fiber coordinate ($y$-axis).}
\label{Boundary_Torus}
\end{figure}

In fact, $p|_{\partial\mathcal{M}}$ lifts to a map $\mathring{p}:\partial\mathcal{M}\to \mathbb{S}^{1}$ which distinguishes each patch's dominant flow direction; the fibers of this map are homeomorphic to $\mathbb{S}^{1}$, and $(\partial\mathcal{M},\mathring{p},\mathbb{S}^{1})$ is the trivial circle bundle over $\mathbb{S}^{1}$. As evidence for the continuum model described above, we computed the lifted flow directions of the patches from the 28 step edge circles $X_{\text{step}}\subset X(50,60)$ we identified, roughly producing a discrete approximate circle bundle $(X_{\text{step}}, \mathring{p},\mathbb{S}^{1})$ in the sense of~\cite{turow2025discrete}. Using the same pipeline outlined in Section~\ref{sec: Extended Model Evidence}, we confirmed the global triviality of the bundle, then constructed a global toroidal coordinate system for $X_{\text{step}}$; Figure~\ref{Boundary_Torus} shows a sample of coordinatized patches which clearly reflects the underlying structure of  $\partial \mathcal{M}$. \\

It would be interesting to investigate whether the full continuous manifold structure is easily detectable for patches of larger size -- the potential value of such a model of fixed low dimension also increases dramatically for spaces of larger patches. \\    

\section{Summary And Future Work}\label{sec: Future Work}
Our analyses expand on the geometric models
previously studied in~\cite{Horizontal_Flow_Circle} and~\cite{opt_flow_torus}
for the space of $3 \times 3 $ high-contrast optical flow patches from~\cite{Sintel}. 
In addition to confirming and expanding upon the torus model, we identified a family of circular dense core subsets corresponding to the binary step-edge range patches described in~\cite{Range_Patches}. In fact, we showed that nearly all of the optical flow patches in the top 1 percent by contrast norm are concentrated near these circles, and observed that these exceptional patches anecdotally are found at motion boundaries in the Sintel video. It would be well worth investigating whether these disjoint circles and the optical flow torus give rise to a single connected manifold structure for patches of larger size, as it could form the basis for geometric compression or classification algorithms  similar to those  developed in~\cite{perea2014klein}. \\

Lastly, our 3-manifold model for the dense subset identified by Adams et al.~\cite{opt_flow_torus} provides an explanation for why the torus model could not be clearly detected with a direct persistence computation, despite the compelling indirect evidence obtained from a fiberwise analysis. Indeed, our results highlight that the relationship between local and global persistence computations depends critically on the geometry of the chosen feature map.\\

\renewcommand{\thesection}{A}

\section{Appendix}
\vspace{2mm}
\subsection{Proof Of Proposition 1}\label{sec: Proof Of Proposition}

Here we provide the proof of Proposition 1. For later reference, we first note that $F^{\perp}(\alpha,\theta)$ (Equation~\eqref{eq: torus model}) satisfies the following properties:

\begin{enumerate}\renewcommand{\labelenumi}{\Roman{enumi}.}
    \item $F^{\perp}(\alpha,\theta)\cdot F(\alpha',\theta) = 0$ for all $\alpha'\in\mathbb{S}^{1}$,

    \item $F^{\perp}(\alpha,\theta)\cdot F(\alpha,\theta') = 0$ for all $\theta'\in\mathbb{S}^{1}$,

    \item $F^{\perp}(\alpha,\theta)$ is D-orthonormal to $F(\alpha',\theta)$ for all $\alpha'\in\mathbb{S}^{1}$

    \item $F^{\perp}(\alpha,\theta)$ is D-orthonormal to $F(\alpha,\theta')$ for all $\theta'\in\mathbb{S}^{1}$
\end{enumerate}

\noindent These can be easily checked by direct computation.  

\vspace{1mm}
\begin{enumerate}
    \item The fact that $F(\alpha,\theta)$ has mean 0 immediately implies that $\tilde{F}(r,\alpha,\theta)$ has mean 0 as well (for any $r$). To see that $\tilde{F}(r,\alpha,\theta)$ is contrast-normalized, observe that $\text{cos}(\tau)$ is an increasing function of $r$ with $\text{cos}\left(\tau(0)\right) = \frac{\sqrt{2}}{2}$ and $\text{cos}\left(\tau(1)\right) = 1$. Since $F(\alpha,\theta)$ and $F^{\perp}(\alpha,\theta)$ are D-orthonormal, it follows from the Pythagorean theorem that $\tilde{F}(r,\alpha,\theta)$ has contrast norm 1. \\ 
    
    \noindent To see that $\tilde{F}(r,\alpha,\theta)$ has predominant direction $\theta$ (mod $\pi$) and directional strength $r$, choose any $\alpha,\theta\in\mathbb{S}^{1}$, and let $A,B,M\in M_{9\times 2}$ respectively denote the matrices whose columns are the component vectors of the patches $F(\alpha,\theta),$ $F^{\perp}(\alpha,\theta)$ and $\tilde{F}(r,\alpha,\theta)$.  Evidently, we have   

\begin{equation*}
    M^{T}M = \text{cos}^{2}\left(\tau(r)\right)A^{T}A + \text{sin}^{2}\left(\tau(r)\right)B^{T}B + \text{cos}\left(\tau(r)\right)\text{sin}\left(\tau(r)\right)(B^{T}A + A^{T}B)    
\end{equation*}
        
    Property II of $F^{\perp}(\alpha,\theta)$ immediately implies $B^{T}A = A^{T}B = 0$.  Moreover, the matrices $A^{T}A$ and $B^{T}B$ decompose as

        \begin{equation*}
            A^{T}A = R_{\theta}\begin{pmatrix}
            1 & 0 \\
            0 & 0 
            \end{pmatrix} R_{\theta}^{T}, \hspace{2cm}
            B^{T}B = R_{\theta}\begin{pmatrix}
            0 & 0 \\
            0 & 1 
            \end{pmatrix} R_{\theta}^{T}
        \end{equation*}

        where,
    
        \begin{equation*}
            R_{\theta} = \begin{pmatrix}
            \text{cos}(\theta) & -\text{sin}(\theta) \\
            \text{sin}(\theta) & \text{cos}(\theta) 
            \end{pmatrix} 
        \end{equation*}

    Thus, we have

\begin{equation*}
    M^{T}M = \text{cos}^{2}\left(\tau(r)\right)R_{\theta}\begin{pmatrix}
                1 & 0 \\
                0 & 0 
                \end{pmatrix} R_{\theta}^{T} + \text{sin}^{2}\left(\tau(r)\right)R_{\theta}\begin{pmatrix}
                0 & 0 \\
                0 & 1 
                \end{pmatrix} R_{\theta}^{T}    
\end{equation*}

\begin{equation*}
    = R_{\theta}\begin{pmatrix}
                \text{cos}^{2}\left(\tau(r)\right) & 0 \\
                0 & \text{sin}^{2}\left(\tau(r)\right)
                \end{pmatrix} R_{\theta}^{T}    
\end{equation*}

    which shows that the eigenvalues of $M^{T}M$ are $\text{cos}^{2}\left(\tau(r)\right)$ and $\text{sin}^{2}\left(\tau(r)\right)$. Since $\frac{\sqrt{2}}{2} < \text{cos}\left(\tau (r)\right) \leq 1$, we have $\text{cos}^{2}\left(\tau(r)\right) > \text{sin}^{2}\left(\tau(r)\right)$, so the predominant direction of $\tilde{F}(r,\alpha,\theta)$ is $\theta$ (mod $\pi$) and the directional strength of $\tilde{F}(r,\alpha,\theta)$ is given by

\begin{equation*}
    \frac{\text{cos}^{2}\left(\tau(r)\right) - \text{sin}^{2}\left(\tau(r)\right)}{\text{cos}^{2}\left(\tau(r)\right)}\hspace{1mm}=\hspace{1mm} 2 - \text{sec}^{2}\left(\tau(r)\right) \hspace{1mm}=\hspace{1mm} r
\end{equation*}

    \vspace{5mm}
    \item Observe that $\text{cos}\left(\tau(1)\right) = 1$. 
    
    \vspace{5mm}
    \item Suppose $\tilde{F}(r,\alpha,\theta) = \tilde{F}(r',\alpha',\theta')$.  By Part 1, we have $r' = r$ and $\theta' = \theta$ (mod $\pi$).  Thus, we have

\begin{equation*}
    \text{cos}\left(\tau(r)\right)\left(F(\alpha,\theta) - F(\alpha', \theta + k\pi)\right) + \text{sin}\left(\tau(r)\right)\left(F^{\perp}(\alpha,\theta) - F^{\perp}(\alpha,\theta + k\pi)\right) = 0
\end{equation*}
    
    where $k\in \{0,1\}$. On the other hand, one easily verifies that $F^{\perp}(\alpha', \theta + \pi) = -F^{\perp}(\alpha',\theta)$, so Property 2 of $F^{\perp}$ above implies $F(\alpha,\theta) - F(\alpha', \theta + k\pi)$ is orthogonal to $F^{\perp}(\alpha,\theta) - F^{\perp}(\alpha', \theta + k\pi)$.  Thus, taking the magnitude squared of both sides of the equation above yields
    \begin{equation*}                                    \text{cos}^{2}\left(\tau(r)\right)\left|F(\alpha,\theta) - F(\alpha',\theta + k\pi)\right|^{2} + \text{sin}^{2}\left(\tau(r)\right)\left|F^{\perp}(\alpha,\theta) - F^{\perp}(\alpha',\theta + k\pi)\right|^{2} = 0
    \end{equation*}

    Since $\text{cos}\left(\tau(r)\right) > 0$, we conclude $F(\alpha', \theta + k\pi) = F(\alpha,\theta)$. Finally, we know from \cite{opt_flow_torus} that $F$ is a double cover whose only identifications are $F(\alpha + \pi, \theta + \pi) = F(\alpha,\theta)$, so these are the only identifications for $\tilde{F}$ as well.   
    \vspace{5mm}
    \item Thinking of $\mathcal{T}$ as a submanifold of the 3-sphere $\tilde{S}$ spanned by $\{e_{1}^{u},e_{2}^{u},e_{1}^{v},e_{2}^{v}\}$, the normal space to each point of $\mathcal{T}$ is 1-dimensional. Properties I and II of $F^{\perp}(\alpha,\theta)$ imply that $F^{\perp}(\alpha,\theta)\perp T_{F(\alpha,\theta)}\mathcal{T}$, though one can also check directly that $F^{\perp}(\alpha,\theta)$ is in the cokernel of $F_{*}(\alpha,\theta)$ (identifying $T_{F(\alpha,\theta)}\mathbb{R}^{18}$ with $\mathbb{R}^{18}$ in the usual way). Since $F^{\perp}(\alpha,\theta)$ has contrast norm 1 for all $\alpha,\theta$, we have $F^{\perp}(\alpha,\theta)\in T_{F(\alpha,\theta)}\tilde{S}$, so the proposition follows.
    
\end{enumerate}

\vspace{5mm}
\subsection{Binary Step-Edge Range Image Patches}\label{sec: Step-Edge Patches}

\vspace{5mm}
\begin{figure}[h!]
  \centering
  \includegraphics[width=0.85\linewidth]{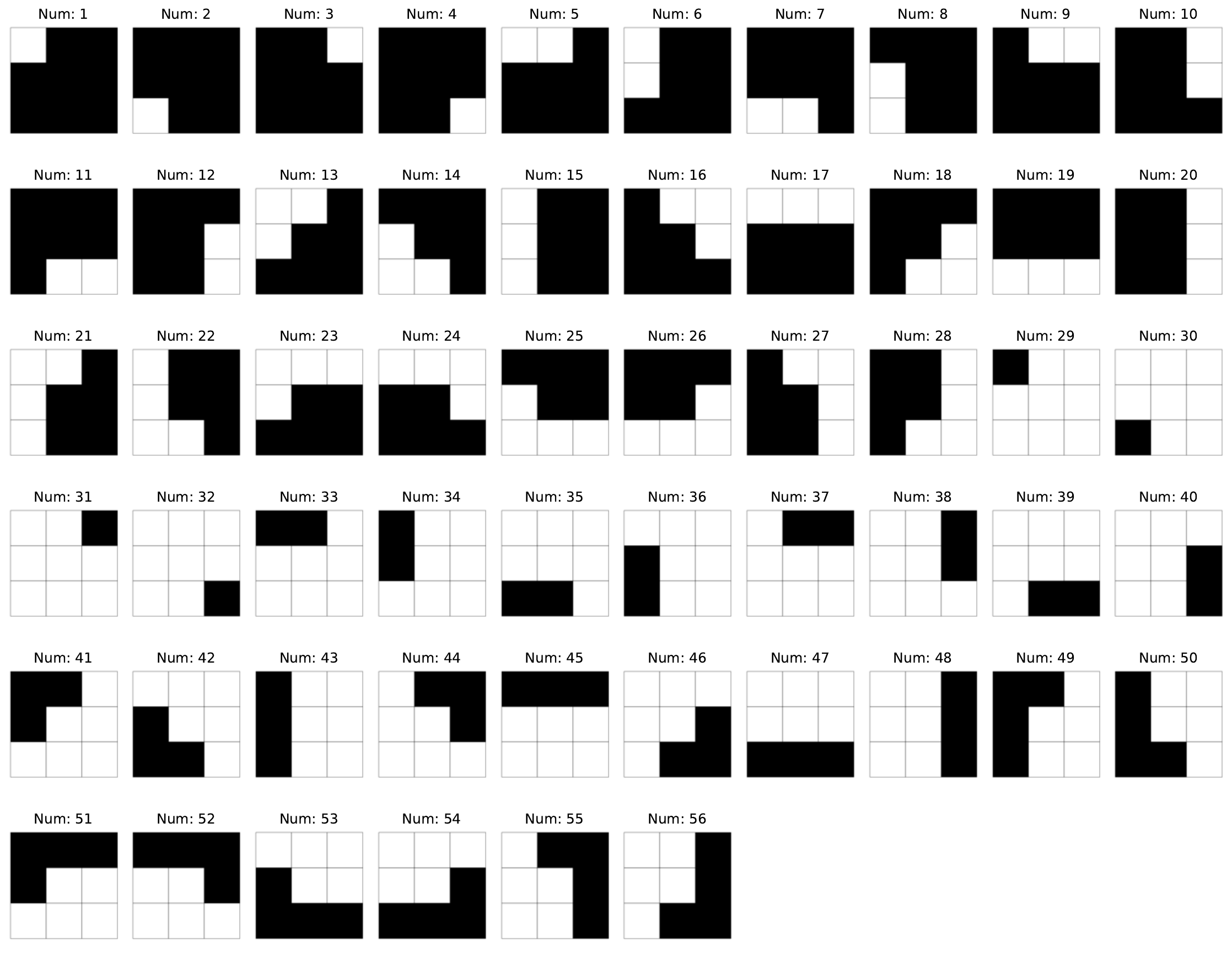}
  \caption{A list of the 56 binary step-edge range image patches}   
  \label{All_Binary_Range_Patches}
\end{figure}

\vspace{2mm}

\printbibliography

\end{document}